\documentclass[lettersize,journal]{IEEEtran}
\usepackage{amsmath,amsfonts}
\usepackage{algorithmic}
\usepackage{algorithm}
\usepackage{array}
\usepackage[caption=false,font=normalsize,labelfont=sf,textfont=sf]{subfig}
\usepackage{textcomp}
\usepackage{stfloats}
\usepackage{url}
\usepackage{verbatim}
\usepackage{graphicx}
\usepackage{cite}
\usepackage{color}

\usepackage{pifont} 

\RequirePackage{xspace}
\makeatletter
\DeclareRobustCommand\onedot{\futurelet\@let@token\@onedot}
\def\@onedot{\ifx\@let@token.\else.\null\fi\xspace}

\usepackage{threeparttable}
\usepackage{booktabs}

\usepackage{hyperref}
\definecolor{urlblue}{rgb}{0.12,0.49,0.85}
\hypersetup{
    colorlinks,
    citecolor=urlblue,
    filecolor=black,
    linkcolor=black,
    urlcolor=urlblue
}

\newcommand{\datasetname}{{\sc SynHead100 }}

\definecolor{myPurple}{rgb}{0.4, .0, .8}
\definecolor{myGreen}{rgb}{0, .8, .3}
\definecolor{myRed}{rgb}{0.8, .2, .2}
\definecolor{myOrange}{rgb}{0.8, 0.45, 0.0}
\definecolor{myBlue}{rgb}{.0, .0, 1.0}

\usepackage{arydshln} 
\usepackage{booktabs}
\usepackage{multirow}

\usepackage{graphicx}
\usepackage{etoolbox}
\usepackage{caption}
\newcommand{\insertfig}{
    \vspace{0.1in}
    \includegraphics[width=\linewidth]{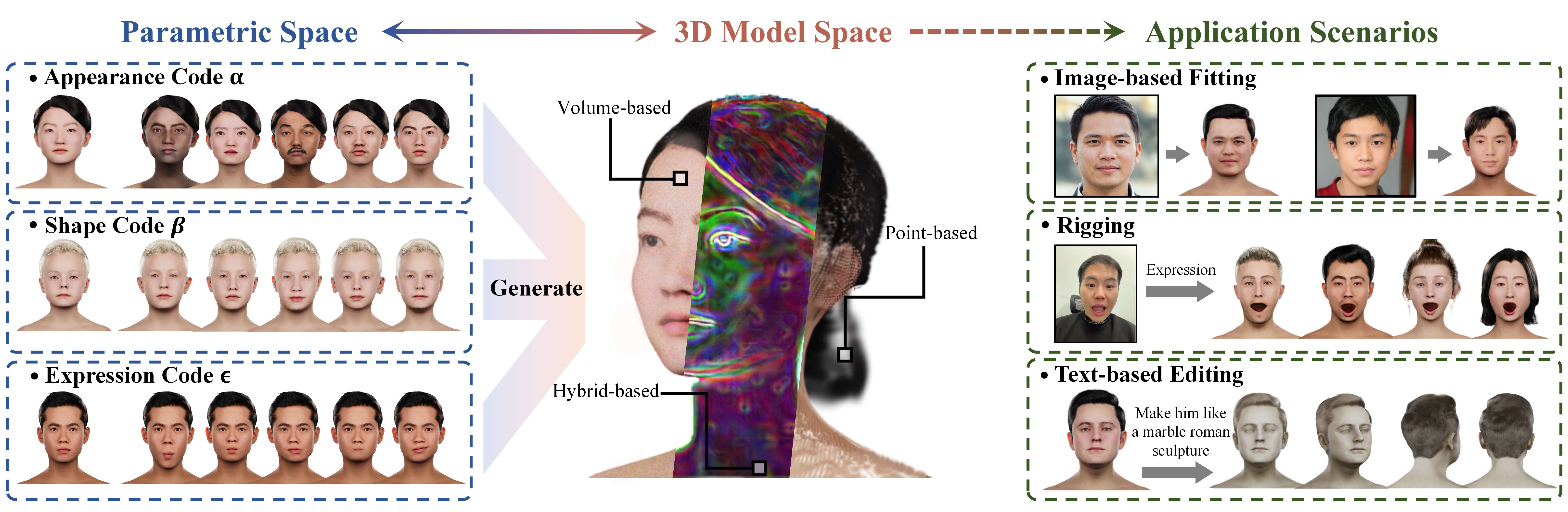}
    \captionof*{figure}{We study learning a native generative head model from limited 3D head datasets.  To this end, a mapping between a parametric space and a 3D model space under novel representations is established with disentanglement of appearance, shape, and expression.  The trained model can then be utilized for extensive applications, including image-based fitting, rigging, and text-based editing.  
    }
    }

\makeatletter
\apptocmd{\@maketitle}{\centering\insertfig}{}{}
\makeatother

\begin{document}

\title{Towards Native Generative Model \\ for 3D Head Avatar}
    \author{Yiyu Zhuang$^{1,2*}$, Yuxiao He$^{1,2*}$, Jiawei Zhang$^{1,2*}$, Yanwen Wang$^{1,2}$, Jiahe Zhu$^{1,2}$, 
    \\ Yao Yao$^{1,2}$, \{Siyu Zhu$^{3}$, Xun Cao$^{1,2}$, Hao Zhu$^{1,2,\#}$\}~\IEEEmembership{Member,~IEEE}
    \\ $^1$ State Key Laboratory for Novel Software Technology, Nanjing University, China\\
    $^2$ Nanjing University, China \quad $^3$ Fudan University, China
    \thanks{* These authors contribute equally.}
    \thanks{\# Corresponding author.}
}



\maketitle

\begin{abstract}

Creating 3D head avatars is a significant yet challenging task for many applicated scenarios. Previous studies have set out to learn 3D human head generative models using massive 2D image data. Although these models are highly generalizable for human appearance, their result models are not 360$^\circ$-renderable, and the predicted 3D geometry is unreliable. Therefore, such results cannot be used in VR, game modeling, and other scenarios that require 360$^\circ$-renderable 3D head models. An intuitive idea is that 3D head models with limited amount but high 3D accuracy are more reliable training data for a high-quality 3D generative model. In this vein, we delve into how to learn a native generative model for 360$^\circ$ full head from a limited 3D head dataset. Specifically, three major problems are studied: 1) how to effectively utilize various representations for generating the 360$^\circ$-renderable human head; 2) how to disentangle the appearance, shape, and motion of human faces to generate a 3D head model that can be edited by appearance and driven by motion; 3) and how to extend the generalization capability of the generative model to support downstream tasks. Comprehensive experiments are conducted to verify the effectiveness of the proposed model. We hope the proposed models and artist-designed dataset can inspire future research on learning native generative 3D head models from limited 3D datasets.

\end{abstract}

\begin{IEEEkeywords}
3D head avatar, native 3D generative model, image-based fitting, text-based editing, facial animation
\end{IEEEkeywords}

\section{Introduction}
\label{sec:intro}

\IEEEPARstart{C}{reating} 3D head avatars is highly demanded in many downstream scenarios, including metaverse, movie production, and digital avatars, thus drawing tremendous attention in computer vision and computer graphics.  Early research to create high-fidelity 3D head modeling evolved into several independent topics, including 3D facial geometry reconstruction, albedo and lighting modeling, hair modeling, face rigging, etc. These algorithms together form a classic 3D head production workflow, which constitutes photo-realistic 3D head modeling. For a long time, this technology has only been used in high-investment industries like the movies but cannot be adopted in most consumer-level products. The key obstacle is that the cost is too high. Specifically, traditional 3D head modeling schemes rely on unavoidable and expensive manual processing and equipment~\cite{debevec2012light}.  In brief, the longstanding challenge is: How can 3D human heads be modeled effectively and economically?

The 3D head generative model appears to be an ideal solution, which can produce plenty of diverse 3D heads from randomly sampled codes or a customized 3D head given a reference image.  According to the source of training data, approaches to training 3D head generative models can be divided into 2D-based methods and 3D-based methods.  
2D-based methods train the model using massive in-the-wild 2D images that are commonly crawled from the Internet~\cite{sun2023next3d, chan2022efficient}. This large number of diverse facial images ensures the generative model's generalization capability to facial appearance. However, such generated 3D heads recover only frontal faces, and the predicted 3D geometry is unreliable. The image fitting of 2D-based generative models also suffers from poor multi-view consistency. Recent studies attempt to use pre-trained 2D large models to supervise the training of 3D head generation models~\cite{wu2023high, zhang2023dreamface}. These methods are still trained using prior knowledge from large-scale 2D image datasets, so they also suffer from limited 3D accuracy.

In this paper, we study 3D-based methods to build a native generative head model, which has received less attention due to limited 3D data.  3D-based methods learn a generative 3D head model from amount-limited but geometric-detailed 3D head datasets, which are reconstructed from real actors or designed by artists, both of which are expensive. Therefore, the scale of the 3D datasets is inevitably small due to the cost constraints, which are expected to persist for a long time. Training on such small-scale 3D head datasets leads to a poor generalization of identities, meaning that the representative space of the generative model is restricted.  To train highly generalizable models using limited 3D data, we propose disentangling appearance, shape, and expression (facial motion) in the parameter space via explicit constraints, resulting in stronger generative performance.  The disentangling of parametric space also enables the generative model to support the independent control of expression and identity, extending the support for applications like appearance editing and animation generation.

We have also explored different representation models to learn an efficient and high-fidelity 3D head generative model. The classic 3D morphable model (3DMM)~\cite{blanz1999morphable} adopts a 3D polygon mesh model to represent human heads, with the key challenge being the difficulty in modeling fine hair. Recently, novel representation methods, including Neural Radiance Fields (NeRF)~\cite{mildenhall2020nerf, mildenhall2021nerf}, Multi-Planes NeRF~\cite{chan2022efficient}, and 3D Gaussian Splatting (3DGS)~\cite{kerbl20233d} models have emerged, exhibiting superior performance in rendering hair, mouth, and eyes. In this paper, native 3D head generative models are learned based on these three representations, with optimized single-image fitting and editing methods. Comprehensive experiments are conducted to verify the effectiveness of these proposed generative models.

This paper is an extended version of two conference papers - MoFaNeRF~\cite{zhuang2022mofanerf} and Head360~\cite{he2022head360}, which learn 3D generative head models with novel representations from 3D head datasets. The new contributions of this paper include achieving a higher-quality rendered 3D face generation model under the 3D Gaussian Splatting representation framework with a customized single-image fitting method. This extension and previous research revolve around one goal: realizing a high-quality 3D head generation model based on a limited yet fine-grained 3D head dataset. Comprehensive experiments are conducted to verify that our proposed models achieve state-of-the-art performance in both 3D generation accuracy and rendering quality. To the best of our knowledge, we present the first native 3D generative model for 360$^\circ$ renderable and riggable 3D heads.

Our contributions are summarized as:
\begin{itemize}
    \item We study approaches to enhancing the generalization capability of the parametric 3D head model by disentangling appearance, shape, and expression in the parameter space. These approaches lead to highly generalizable 3D head models with limited training data.
    \item Novel representation models, including NeRF, Multi-Planes NeRF, and 3DGS, are explored to learn an efficient and high-fidelity 3D head generative model, overcoming challenges in modeling fine details of human heads.
    \item A higher-quality rendered 3D face generation model is achieved under the 3DGS framework with a customized single-image fitting method, extending previous research.
    \item To the best of our knowledge, we present the first native 3D generative model for 360$^\circ$ renderable and riggable 3D heads, with experiments verifying state-of-the-art performance in 3D geometric accuracy and rendering quality on \datasetname dataset.
\end{itemize}
\section{Related Work}
\label{sec:related}

\noindent\textbf{3D Morphable Model (3DMM): }
3DMM is defined as a statistical model that transforms the 3D shape and texture of the heads into a vector space representation\cite{blanz1999morphable}. The parametric transformation is learned from a large set of heads in diverse shapes, appearances, and expressions, represented by registered 3D polygonal mesh models. According to the properties of parametric transformation, 3DMM can be further divided into linear and nonlinear models. For a linear 3DMM, Principal Component Analysis (PCA)~\cite{pearson1901liii} is commonly used to learn the linear mapping from the 3D vertices and texture space of the head model to a low-dimensional space~\cite{vlasic2005face, cao2013facewarehouse, li2017learning, jiang2019disentangled, yang2020facescape, zhu2023facescape, JNR}. Linear 3DMM conversion calculation is straightforward and is widely used in face alignment, face reconstruction, and other tasks. For a non-linear 3DMM, a neural network or other non-linear model is adopted to model the mapping~\cite{bagautdinov2018modeling, tewari2018self, tran2019learning, tran2018nonlinear, cheng2019meshgan, tran2019towards}, which is more powerful in representing detailed shape and appearance than linear mapping.
3DMM is widely used in 3D face reconstruction~\cite{zhu2017face, zhu2016face, xiao2022detailed}, generation~\cite{zhuang2022mofanerf, hong2022headnerf, wu2023high}, talking faces~\cite{yu2022migrating, sun2023vividtalk, He2022Emo3DTalk, chen2024emotion}, and other downstream applications. 
The main problem with 3DMM is that it struggles to model hair, beards, and eyelashes, as 3D polygon mesh models can hardly represent these complex and detailed 3D structures. 
We recommend referring to the recent survey\cite{egger20203d} for a comprehensive review of 3DMM.

\begin{figure*}[th]
    \centering
    \includegraphics[width=1.0\linewidth]{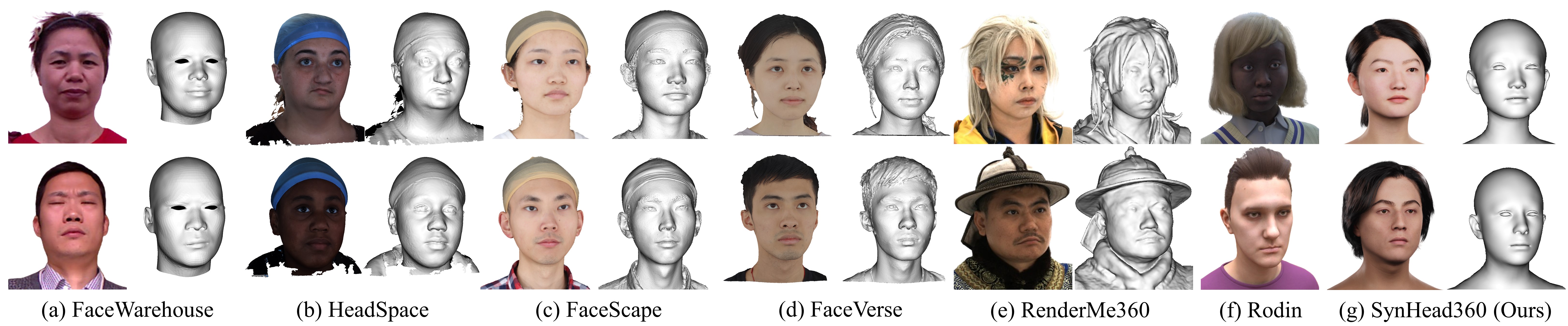}
    \caption{\textbf{Visualization of 3D head datasets.}
    }
    \label{fig: datasets}
\end{figure*}

\noindent\textbf{Face and Head Datasets: }
Datasets for training 3D head generation models can be classified into 2D and 3D datasets. 2D datasets are commonly collected from the internet (\textit{e.g.} FFHQ\cite{karras2019style}), consisting of millions of 2D portrait images covering a wide range of personal appearances and background environments~\cite{liu2015faceattributes, karras2019style, lee2020maskgan}.
The main disadvantage of 2D datasets is the lack of ground-truth 3D models corresponding to 2D images, which limits the geometric accuracy of the 3D head generative models trained on such datasets. As another data source, 3D datasets are obtained through 3D reconstruction systems or artist designs, providing 2D images and corresponding reliable 3D head models~\cite{zhu2023facescape, wang2022faceverse, dai2020statistical}.  The precise models in 3D datasets bring accurate detail precision to 3D head generative models. However, the quantity of 3D datasets is usually much smaller than that of 2D datasets due to the cost of the 3D capturing system and labor, bringing a more significant challenge to training a well-generalized model. 
Regarding 3D head datasets, previous multi-view or 3D head datasets exhibit notable shortcomings, such as low 3D accuracy~\cite{cao2013facewarehouse}, wrapped hair~\cite{yang2020facescape, dai2020statistical, wang2022faceverse, zhu2023facescape}, and lack of rigging~\cite{dai2020statistical, pan2023renderme}, as shown in Fig.~\ref{fig: datasets} and Tab.~\ref{tab:data}. Recently, learning from large-scale synthetic 3D heads~\cite{wood2021fake} has proven effective~\cite{wang2023rodin}. However, their synthetic 3D head dataset is not publicly available and incurs high costs.  Consequently, we propose to create a high-fidelity synthetic 3D head dataset for free research use, containing $100$ distinct subjects with diverse hairstyles and facial appearances, rigged into $52$ standard blendshape bases. The models were designed by artists, referencing FaceScape~\cite{zhu2023facescape} and HeadSpace~\cite{dai2020statistical}, two public datasets that cover diverse identities worldwide.

\noindent\textbf{2D Generative Head Model: }
Research on deep generative models has made significant progress in the last decade. The conditional generative models represented by the generative adversarial network (GAN)~\cite{goodfellow2020generative} have produced excellent results in the task of 2D face generation~\cite{karras2019style, karras2020analyzing, deng2020disentangled, deng2022gram, chan2021pi}. These methods learn a distribution of a large-scale 2D face image dataset through a deep generative network, which is trained by a min-max game between the generator network and the discriminator network. GAN-based models can synthesize 2D faces given a parameter and obtain the parameters that fit a given 2D face by GAN inversion methods~\cite{xia2022gan}.  GAN-based models are commonly trained on many real-captured portraits, so their generated faces are highly photo-realistic. The disadvantage of GAN-based models is that they lack 3D information, so large-angle view synthesis is not supported. We highly recommend reading the survey on GANs~\cite{creswell2018generative, gui2021review} and GANs for face~\cite{kammoun2022generative} for a comprehensive understanding of this field.  Very recently, diffusion models~\cite{ho2020denoising} for face generation have shown appealing performance in cross-modal face generation~\cite{kim2023dcface, huang2023collaborative, lan2023gaussian3diff}. However, similar to GAN-based generative models, they also suffer from the lack of stable 3D information. Moreover, the diffusion-based models have slow inference speed due to iterative denoising process.

\noindent\textbf{Generative Head NeRF}
NeRF~\cite{mildenhall2020nerf, mildenhall2021nerf} provides an alternative representation for the parametric head by combining an implicit neural network with a volume renderer. 
A straightforward idea is to combine NeRF with GAN to create a conditional 3D GAN~\cite{toshpulatov2021generative}. Early works~\cite{niemeyer2021giraffe, gu2021stylenerf, chan2022efficient, sun2023next3d, sun2022fenerf} train generative 3D GANs by using massive 2D real portraits and are surprised to find that the model can learn 3D information from these 2D portraits with various identities and views. However, the learned 3D information is limited and can be rendered only at small angles in the front, while larger angles will lead to severe rendering degradation. HeadNeRF~\cite{hong2022headnerf} proposes to train the model with both large-scale 2D head images and relatively few 3D models, but only extending the renderable view to about $60^\circ$. MoFaNeRF~\cite{zhuang2022mofanerf} only leverages high-quality multi-view head images for training and realizes a $180^\circ$ free-view renderable parametric head model. However, the main problem is that existing high-quality multi-view 3D head data are commonly captured with hair tied up, so MoFaNeRF does not model hair. In later studies, RODIN~\cite{wang2023rodin} trains the parametric head with large-scale multi-view images rendered from artist-designed 3D models, and PanoHead~\cite{an2023panohead} combined large-scale frontal 2D images with captured back-view hair images for training, both of which achieved a $360^\circ$ renderable parametric head model. The main problem with these two models is that their facial expressions and motions are not riggable. 

\noindent\textbf{Generative 3D Gaussian Splats}
Recent developments in 3DGS representation ~\cite{kerbl20233d, chen2024survey} have shown significant improvements in novel view synthesis compared to NeRF. This method employs a collection of Gaussian ellipsoids to depict scenes, enabling real-time rendering through rasterization. To leverage its efficient rendering capabilities, many studies have extended its application to tasks such as 3D head reconstruction ~\cite{xu2024gaussian, wang2023gaussianhead} and 3D full-body avatars reconstruction \cite{hu2024gauhuman, hu2024gaussianavatar, li2024animatable}. These pipelines are designed to generate detailed and animatable avatars from videos, necessitating scene-specific optimization.
Another line of research focuses on the generalization of 3D avatar based on 3DGS, including text-driven synthesis using Score Distillation Sampling ~\cite{huang2024humannorm, tang2023dreamgaussian, ling2024align, ren2023dreamgaussian4d} and integration with GAN ~\cite{kirschstein2024gghead, abdal2024gaussian}. However, text-driven synthesis methods are limited by their extensive computational requirement, often taking hours to optimize a single scene.
GGHead ~\cite{kirschstein2024gghead} and Gaussian Shell Map ~\cite{abdal2024gaussian} introduce a Gaussian-based 3D GAN. However, the Gaussian Shell Map is constructed using multiple shells derived from the parametric template, which strongly relies on the template model for animating the generated outputs. Consequently, this method cannot effectively model facial expressions or capture variations caused by different poses. On the other hand, GGHead represents the head with a template mesh and explicit UV-warping attributes with learnable offset, but it lacks control over facial expressions and the capability to synthesize faces at large angles.

Unlike all the above models, our model is the first motion-riggable and $360^\circ$ renderable high-fidelity parametric head model.

\begin{table*}[]
\centering
\caption{\textbf{Comparisons of 3D head datasets.}}
\begin{threeparttable}
\begin{tabular}{@{}lcccccllccccc@{}}
    \cmidrule(r){1-6} \cmidrule(l){8-13}
    \multicolumn{1}{c}{Dataset}                                & Sub. Num. & Range & Hair                       & Rig                        & \multicolumn{1}{l}{Source} &   & \multicolumn{1}{c}{Dataset}                               & \multicolumn{1}{c}{Sub. Num.} & \multicolumn{1}{c}{Range} & \multicolumn{1}{c}{Hair}           & \multicolumn{1}{c}{Rig}    & Source                      \\ \cmidrule(r){1-6} \cmidrule(l){8-13} 
    BU-3DFE~\cite{yin20063d}             & 100  & front & \ding{55} & \ding{51} & Active                     &  & BP4D-S~\cite{zhang2014bp4d}         & 41                       & front & \ding{51}         & \ding{51} & Passive \\
    BU-4DFE~\cite{zhang2013high}         & 101  & front & \ding{51} & \ding{55} & Active                     &  & HeadSpace~\cite{dai2020statistical} & 1519                     & $270^{\circ}$             & \ding{55}         & \ding{55} & Active                      \\
    BJUT-3D~\cite{baocai2009bjut}        & 500  & front & \ding{55} & \ding{55} & Active                     &  & FaceScape~\cite{zhu2023facescape}   & 938                      & $360^{\circ}$             & \ding{55}         & \ding{51} & Passive                     \\
    Bosphorus~\cite{savran2008bosphorus} & 105  & front & \ding{55} & \ding{51} & Active                     &  & FaceVerse$^\dagger$~\cite{wang2022faceverse}  & 128                      & $360^{\circ}$             & \ding{51} & \ding{51} & Hybrid                      \\
    FaceWarhouse\cite{cao2013facewarehouse}       & 150  & front & \ding{51} & \ding{51} & Active                     &  & Rodin$^\ast$~\cite{wang2023rodin}   & $10^5$            & $360^{\circ}$             & \ding{51}         & \ding{55} & Manual                      \\
    4DFAB~\cite{cheng20184dfab}          & 180  & front & \ding{55} & \ding{55} & Hybrid                     &  & RenderMe360$^\dagger$~\cite{pan2023renderme}  & 500                      & $360^{\circ}$             & \ding{51} & \ding{55} & Passive                     \\
    D3DFACS~\cite{cosker2011facs}        & 10   & front & \ding{51} & \ding{51} & Passive                    &  & SynHead100(Ours)                        & 100                      & $360^{\circ}$             & \ding{51}         & \ding{51} & Manual                      \\ \cmidrule(r){1-6} \cmidrule(lr){8-13}
\end{tabular}

\begin{tablenotes}
    \footnotesize
    \item[$\ast$] This dataset is not publicly available; \quad $\dagger$  The captured 3D hair shape in this dataset is rough .
  \end{tablenotes}
\end{threeparttable}

\label{tab:data}
\end{table*}

\section{Generative 3D Head Models}
\label{sec: models}

This paper aims to learn a native generative model for 3D heads that achieves random generation, single-image fitting, animating, and editing. This section first introduces the core model design, including the representations for generative 3D head models (Sec.~\ref{sec: rep}) and our design to map the parametric space to 3D head representative space (Sec.~\ref{sec: param}). Then volume-based model (Sec.~\ref{sec: model_vol}), hybrid-based model (Sec.~\ref{sec: model_hyb}) and point-based model (Sec.~\ref{sec: model_pt}) are introduced, respectively.

Before we dive deeper, we would like to clarify the \textit{generative model} we stated and the current generative models in deep learning. Current generative models in deep learning aim at estimating a distribution $p(x)$ given a dataset $x\sim X$. We can explicitly marginalize $p(x)$ via latent variable $z$:
\begin{equation}
    \begin{split}
        p\left( x \right) =\int{p\left( x|z \right) p\left( z \right) \mathrm{d}z}
    \end{split}
\end{equation}
The general approach is to sample $z$ from a Gaussian or uniform distribution and learn the underlying distribution from a large amount of data. While utilizing unordered 2D image datasets, some characteristics (\textit{e.g.} glass, smile) can be unsupervised decoupled. It does not lead to accurate 3D geometric information, resulting in multi-view inconsistency and identity change.

One straightforward approach to tackling this issue is by leveraging well-structured 3D datasets.  Although 3D datasets provide clear decoupling (e.g., expression bases, shape coefficients) and precise geometries, they face the challenge of capturing large-scale identities, as previously mentioned.  In this context, $p(z)$ essentially degenerates into an approximation that yields a finite number of Dirac distributions for $z_{id}$.    Consequently, our objective shifts toward maximizing the generalizability of $p(x|z_{id})$ on a limited dataset by refining the design of the decoupled parameter space.    We believe such conditional probabilistic models can be equally generative when viewed from the perspective - with sufficient interpolation capacity and a well-defined parameter space.

\subsection{3D Representations} 
\label{sec: rep}

Representing 3D head models is a fundamental problem for native 3D generative models so that we will start with their representations.  Traditional 3D generative head models, such as 3DMM~\cite{blanz1999morphable}, adopt 3D polygon mesh models to represent 3D human heads. Though 3D polygon mesh models are efficiently structured, they struggle to model complex geometry and materials, such as hair. In pursuit of high-quality rendering within a unified 3D representation, we explore using novel neural representations to establish 3D generative head models, including \textit{volume-based models, hybrid-based models, and point-based models}.

\noindent\textbf{Volume-based Representation.} 
Volume-based models represent 3D scenes by capturing volumetric geometry and appearance through implicit functions.
The representative method is Neural Radiance Field (NeRF)~\cite{mildenhall2020nerf, mildenhall2021nerf}, which achieves high-fidelity free-view rendering of 3D scenes by training on multi-view images. 
In this paper, we first study how to build a conditional generative model of 3D human heads based on NeRF representation.
It represents the radiance field as volumetric density $\sigma$ and color $\mathbf{c}=(R, G, B)$. An MLP is used to predict $\sigma$ and $\mathbf{c}$ from a 3D point $\mathbf{x}=(x,y,z)$ and viewing direction $\mathbf{d}=(\theta,\phi)$. We follow vanilla NeRF and introduce position encoding to transform the continuous inputs $\mathbf{x}$ and $\mathbf{d}$ into a high-dimensional space.
The field of $\sigma$ and $\mathbf{c}$ can be rendered to images using a differentiable volume rendering module. For a pixel in the posed image, a ray $\mathbf{r}$ is cast through the neural volume field from the ray origin $\mathbf{o}$ along the ray direction $\mathbf{d}$ according to the camera parameters. 
This process is formulated as $\mathbf{r}(z)=\mathbf{o}+z\mathbf{d}$, where $z$ denotes the distance along the ray.
To calculate a pixel's output color $\mathbf{C(r)}$, points along the ray are sampled and evaluated using the MLP. Their density $\sigma(\cdot)$ and RGB values $\mathbf{c}(\cdot)$ are then accumulated by $\mathcal{F}$, which is formulated as:

\begin{equation}
    \begin{split}
        \mathbf{C(r)}\!=\!\int_{z_{n}}^{z_{f}} T(z) \sigma( \mathbf{r}(z)) \mathbf{c}(\mathbf{r}(z),\mathbf{d})dz\,,\\
         \:\textup{where\ }\:T(z)\!=\!\exp\!\left(-\int_{z_{n}}^{z}\sigma (\mathbf{r}(s))\mathrm{ds}\right) .
    \end{split}
\end{equation}

\noindent where $T(t)$ is defined as the accumulated transmittance along the ray from $z_n$ to $z$, where $z_n$ and $z_f$ are near and far bounds.  
The MLP's parameters can be learned by minimizing a photometric loss between the rendered color and ground truth.  In this way, a single 3D human head is represented by an implicit function predicting a volume-based model.

\begin{figure}[th]
    \centering
    \includegraphics[width=1.0\linewidth]{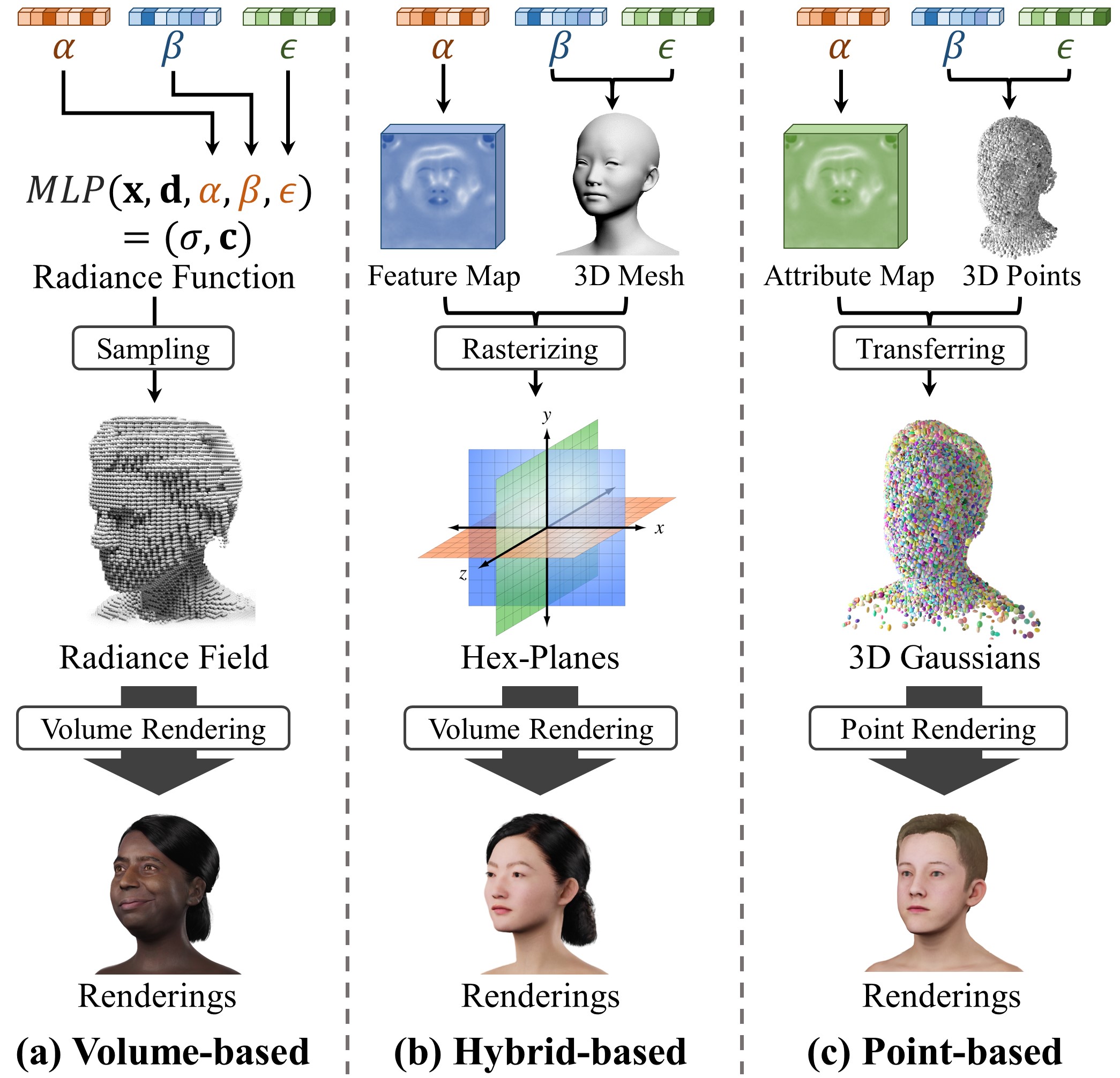}
    \caption{\textbf{Illustration of volume-based, hybrid-based, and point-based 3D head representations.} (a) Volume-based representation employs MLP for representing the radiance field, resulting in a slower rendering process when generating images through volume rendering. (b) Hybrid-based representation integrates explicit parametric mesh and hex-plane structures with compact MLP, facilitating enhanced control over expressions and increased detail. (c) Point-based representation retains the explicit parametric mesh while incorporating more efficient UV-mapped Gaussian attributes for avatar representation, improving quality and efficiency. 
    }
    \label{fig: rep}
\end{figure}

Based on such a volume-based framework, we extend the input parameters of the implicit function to include $(\alpha, \beta, \epsilon)$ parameters describing the appearance, shape, and motion, respectively. The modified model can represent multiple 3D heads within a single implicit function, thus forming a volume-based generative 3D head model. As shown in Fig.~\ref{fig: rep}-(a), the volume-based generative model is conditioned on parametric space $(\alpha, \beta, \epsilon)$, which are concatenated with points' position $\mathbf{x}$ and viewing directions $\mathbf{d}$ for the MLP to model radiance field for human head. The design of this volume-based model will be detailed in Sec.~\ref{sec: param}.

\noindent\textbf{Hybrid-based Representation.} 
Volume-based models perform well in rendering human head models and outperform mesh-based models in modeling fine hair.  However, the volume-based model struggles with the head animation, resulting in blurred renderings. We consider it due to the inefficiency of implicit functions based on nonlinear neural networks when modeling dynamic geometric structures. Based on such observation, we introduce the hybrid-based model, which represents a NeRF with generative multi-planes~\cite{chan2022efficient} and is structured as a neural texture~\cite{thies2019deferred} attached to a parametric head mesh model~\cite{sun2023next3d}. Specifically, we follow EG3D~\cite{chan2022efficient} to provide explicit volumetric information by compressing neural radiance field into three orthogonal planes. These planes are generated using a 2D Style-GAN~\cite{karras2020analyzing}, which functions to transform noise $z$ into a 3D head.  In the rendering phase, features of the orthogonal planes are gathered at the projected coordinates of sampled points and decoded into volumetric density and color using a lightweight MLP.  This approach results in faster rendering speeds and reduced memory usage compared to the deep MLPs employed in volume-based methods.  Inspired by Next3D~\cite{sun2023next3d}, the neural texture attached to the FLAME model is rasterized to the orthogonal planes to cover the facial region. So, the facial expression is modeled via a linear parametric model, which shows superior performance in expression-rigged animation compared to the volume-based model.

As shown in Fig. \ref{fig: rep}-(b), the hybrid-based model is a hybrid framework involving neural texture, parametric 3D mesh, hex-planes and volume-renderer. To improve the model's capability to represent 360$^\circ$-renderable head, the tri-planes are modified to hex-planes. Additionally, we introduce a two-branch network to model bald heads and hair separately. Our proposed training strategy decouples hair and head, enabling free-swapping of hairstyles and improved fitting performance.

\noindent\textbf{Point-based Representation.} 
Very recently, point-based 3D representation~\cite{kerbl20233d} has attracted widespread attention from researchers for its efficient and high-quality rendering performance. Point-based models explicitly represent the 3D human head as attributed point clouds, each point of which is represented by an anisotropic Gaussian kernel:

\begin{equation}
    \mathcal{G} _i=\left\{ \mu ,\mathbf{s},\mathbf{q},o,c \right\},
\end{equation}

\noindent where each kernel $\mathcal{G}$ is assigned five attributes: position $\mu\in\mathbb{R}^3$, scale $\mathbf{s} \in \mathbb{R}^3$, rotation $\mathbf{q} \in \mathbb{R}^4$ parameterized as quaternions, opacity $o \in \mathbb{R}$, color $c \in \mathbb{R}^3$. A 3D Gaussian is defined by position $\mu$ and a covariance matrix $\Sigma$ by the following formulation:

\begin{align}
    G\left( x \right) &=e^{-\frac{1}{2}\left( x-\mu \right) ^T\Sigma ^{-1}\left( x-\mu \right)},\\
    \Sigma &=RSS^TR^T,
\end{align}

\noindent where the covariance matrix $\Sigma$ is constructed by a rotation matrix $R$ transformed from $\mathbf{q} \in \mathbb{R}^4$ and a scaling matrix $S$ represented by $\mathbf{s} \in \mathbb{R}^3$.
Then, the differentiable tile-based rasterizer can efficiently render the Gaussians into an image by the following equation:

\begin{equation}
    C=\sum_{i=1}{\alpha _i\prod_{j=1}^{i-1}{\left( 1-\alpha _j \right)}c_i},
\end{equation}

\noindent where weight $\alpha$ is determined by the the product of the 2D projection of the 3D Gaussian and a per-point opacity $o \in \mathbb{R}$.
In contrast to volume-based and hybrid-based models, point-based models provide fully explicit control, and primitives bound to the mesh can move freely under the drive of the parametric head. A plausible method is to utilize a powerful grid-based CNN to generate attribute maps with different Gaussian attributes, then sample from those attribute maps via parametric head UV layout as shown in Fig.~\ref{fig: rep}-(c). This strategy has also been adopted in previous works, such as GGHead~\cite{kirschstein2024gghead} and GSM~\cite{abdal2024gaussian}, and will be detailed in Sec.~\ref{sec: model_pt}.

Through experiments, we observed that the point-based model could achieve rendering quality comparable to the hybrid-based model. Meanwhile, the point-based model has a considerable speed advantage and higher expression-animated rendering quality.  The point-based model struggles to generate parts away from the head, such as shoulders and clothing, where 3D points are not well-aligned. We believe it is because the geometric structure of the point-based 3D representation is spatially discrete, unlike the continuous one of the volume-based representation. Generating discrete 3D points is more challenging than generating a continuous 3D volume. We relieve this problem by modeling the hair and head separately while synthesizing the rendered results with a StyleUNet~\cite{wang2023styleavatar}.

\begin{figure*}[th]
    \centering
    \includegraphics[width=1.0\linewidth]{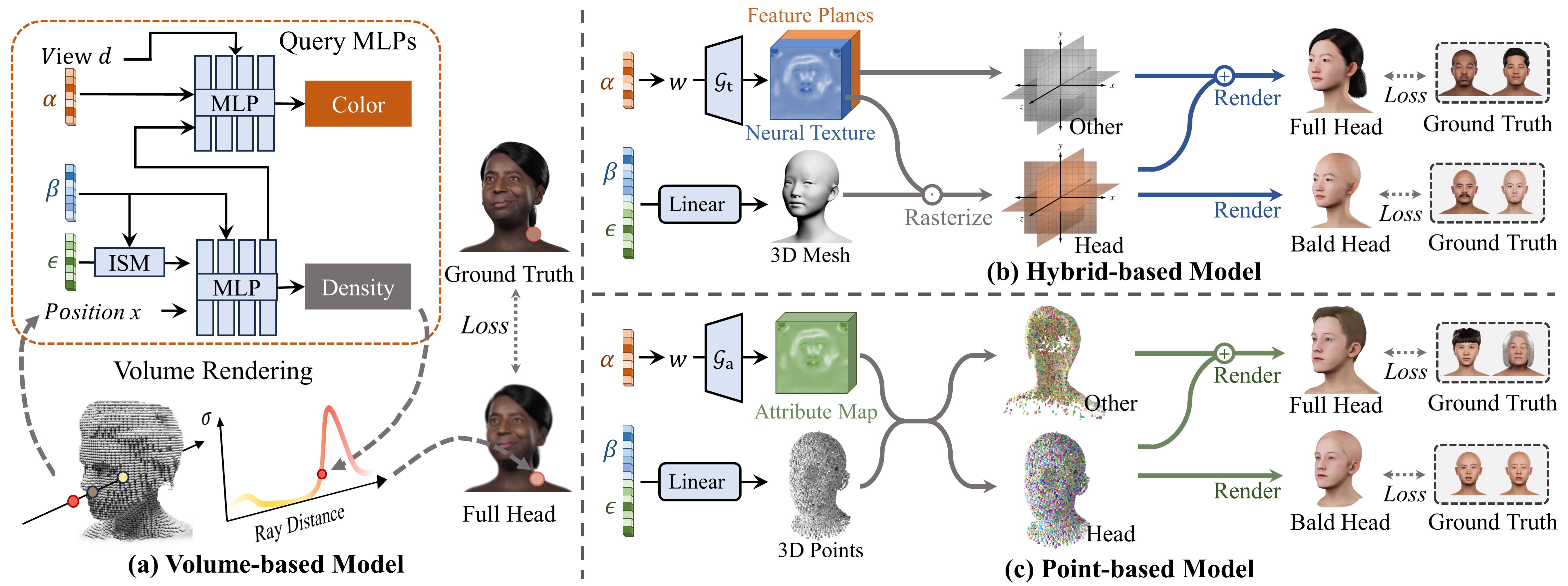}
    \caption{\textbf{The pipelines of volume-based, hybrid-based, and point-based 3D head representations.}  \textbf{(a)} The volume-based model is built by taking the parameter $(\alpha, \beta, \epsilon)$ as conditions of the NeRF function.  Then, the volume rendering is leveraged to transform the radiance field into into images.
    \textbf{(b)} The hybrid-based model introduces neural texture to combine explicit parametric 3D models with implicit neural radiance field. Furthermore, hex-planes are used to model the head and hair separately. This way, the parameters $(\alpha, \beta, \epsilon)$ are disentangled at the model level. 
    \textbf{(c)} The point-based model inherits the idea of the disentanglement of parameters $(\alpha, \beta, \epsilon)$ at the model level. The key difference is to leverage attribute maps instead of neural texture and a point-based rendering process. 
    }
    \label{fig: pipeline}
\end{figure*}
\subsection{Parametric Mapping}
\label{sec: param}

After the representations of the 3D generative model are established, the subsequent question is how to learn an accurate and well-disentangled mapping from the parametric space to the 3D head model space.  Before introducing our solutions, it is necessary to clarify the difference between \textit{`Learning from 2D images'} and \textit{`Learning from 3D models'}.

\textit{Learning from 2D images.} Early extensive work focused on constructing 3D head generative models from millions of unstructured 2D face images.  As no labels about the facial identity and expressions are provided by the unstructured 2D face images, these methods typically directly map the human head images to an unconstrained parametric space. The extensive and diverse training images ensure the trained generative model generalizes facial appearance well.  However, these generative models are not fully 3D -- their rendered results are good only at frontal views within a certain angle, and the learned 3D structure is merely plausible, not precise.

\textit{Learning from 3D models} is an entirely different matter.  3D models are difficult to acquire, so available 3D head models for training are far fewer than unstructured 2D head images.  Despite being disadvantaged in quantity, 3D datasets can provide more precise 3D geometry and more comprehensive annotations, such as standardized facial expressions and disentangled encodings of shape and appearance. We will review 3D datasets and introduce our \datasetname dataset in Sec.~\ref{sec: data}.
Based on observations of 3D datasets, we propose mapping 3D models to a semantically constrained parametric space rather than an unconstrained one. In this predefined parametric space, dimensions of appearance, shape, and expression are disentangled via supervised learning. This allows us to learn highly generalized generative models even with a limited number of 3D head data.

Specifically, we parameterize the 3D head models into shape $\beta$, appearance $\alpha$, and expression $\epsilon$.  Shape code $\beta$ and appearance code $\alpha$ are geometric and photometric features of a human head, respectively, which reflect a person's unique identity.  Expression code $\epsilon$ depicts head states at all possible motions, essentially composed of the movements of the joints and muscles of the face.  By adjusting shape and appearance parameters, a person's identity can be edited, and by adjusting motion parameters, the head can be animated with various expressions or driven by the portrait's captured sequences. Therefore, mapping and disentangling shape, appearance, and motions is crucial for a 3D human head generative model. The accuracy of mapping determines the rendering quality of the final generated model, while the precision of decoupling determines the quality of facial animation synthesis and facial editing.

Overall, we use prior features of shape $\beta$, appearance $\alpha$, and expression $\epsilon$ as conditions for training, enabling the conditional generative model to learn an accurate mapping from the parametric space to the 3D head model space. 
As shown in Tab.~\ref{tab:models} and Fig.~\ref{fig: rep}, specific training strategies and network structures are designed for different 3D representations, including volume-based model (Sec.~\ref{sec: model_vol}), hybrid-based model (Sec.~\ref{sec: model_hyb}), and point-based model (Sec.~\ref{sec: model_pt}).

\subsection{Volume-based Model}
\label{sec: model_vol}

In the volume-based model, we follow NeRF to use a MLP to fit an implicit function representing 3D heads, so $\beta$, $\alpha$, and $\epsilon$ are all directly conditioned on the MLP, formulated as:

\begin{equation}
    \{\sigma, \mathbf{c}\} = MLP(\mathbf{x}, \mathbf{d}, \alpha, \beta, \epsilon).
\end{equation}

A straightforward training strategy is to leverage one-hot encoding to parameterize $\beta$, $\alpha$, and $\epsilon$. Still, it suffers from redundant parameters because the similarity of large-amount heads is repeatedly expressed in one-hot code.  Therefore, we propose a customized training strategy for the three featured parameters in the \textbf{volume-based model}:

\textit{$\bullet$ Shape code $\beta$.} In the training of the volume-based model, we leverage the identity parameter of the bilinear model as training input, which is the PCA factor of the 3D mesh for each subject, as the numerical variation of the identity parameter reflects the similarity between face shapes, which makes the solution space of facial shapes more efficient.

\textit{$\bullet$ Appearance code $\alpha$}. Encoded features of the UV texture maps are leveraged as training input, considering that the UV texture is the ideal carrier to convey the appearance in a spatial-aligned UV space. The texture encoding module (TEM), a CNN-based encoder network is proposed, to transfer the coded appearance information into the MLP. TEM is only used in the training phase, and we find it significantly improves the quality of synthesized images. The reason is that the appearance details are spatial-aligned and are well disentangled from the head shape, relieving the burden of memorizing appearances for the MLP.

\textit{$\bullet$ Expression code $\epsilon$.} learnable code is leveraged as training input. Previous methods\cite{park2021hypernerf, tretschk2021non} try to model the dynamic face by adding a warping vector to the position code $\textbf{x}$, namely deformable volume. We find directly concatenating expression parameters with the position code as \cite{li2022neural, gafni2021dynamic} causes fewer artifacts, and our identity-specific modulation (detailed in Section~\ref{sec: model_vol}) further enhances the representation ability of expression. 

The overall architecture of the volume-based model is illustrated in Fig.~\ref{fig: pipeline}-(a).  Through experiments, we find that the above model cannot render pore-level appearance details, so RefineNet is introduced to synthesize small-scale details based on the generated results.  We adopt Pix2PixHD~\cite{wang2018high} as the backbone of the RefineNet, with the rendered images from the volume-based generative model as input. In the training phase, the overall network, except for RefineNet, is first trained, and then RefineNet is trained with the other parts of the network detached. The loss function and hyper-parameters for training RefineNet are the same as Pix2PixHD. The training details will be explained in Sec.~\ref{sec: train}.

\subsection{Hybrid-based Model}
\label{sec: model_hyb}

Although the volume-based model performs well in rendering quality, it suffers from artifacts in the expression-driven results.  We believe this is because the disentangling capability of MLP-based implicit functions is not sufficiently robust.  Based on this insight, we propose a hybrid-based model that separates appearance features from shape and motion features at the model representation level.  The hybrid-based model is formulated as:

\begin{equation}
    \{\sigma, \mathbf{c}\} = \mathbb{S}_v(x, d, \mathbb{R}(B(\beta, \epsilon), \mathcal{G}_t(\alpha))),
\end{equation}

\noindent where $B$ is the bilinear model generated from the dataset. $\mathcal{G}_t$ is a 2D generative model synthesizing a neural texture from latent code $w$, which is transformed from a learnable code $\alpha$ with an MLP.  $\mathbb{R}$ is a rasterizer that transforms the neural texture into hex-planes; $\mathbb{S}_v$ is a sampler that extracts and decodes radiance value from the hex-planes given $\mathbf{x}$ and $\mathbf{d}$. 
The training strategies of $\beta$, $\alpha$, and $\epsilon$ in the \textbf{hybrid-based model} are explained as:

\textit{$\bullet$ Shape code $\beta$.} A parameterized bilinear model is introduced as the shape encoder. Though the volume-based model also leverages the bilinear model, it uses parameters of the bilinear model as the input of the implicit function to regularize the training. By contrast, the hybrid-based model directly introduces the bilinear model into the generative model, so the follow-up neural network does not need to fit the mapping from $\beta$ to head shapes again. Instead, it can focus on learning the mapping from $\alpha$ to head appearance, achieving better rendering quality.

\textit{$\bullet$ Appearance code $\alpha$.} A neural texture is syntheszied by a 2D generative model conditioned on $\alpha$. Following Next3D~\cite{sun2023next3d}, the neural textures with the generated 3D mesh are rasterized into multiple feature planes and rendered to the image given the camera parameters. The following RefineNet further improves the details of the rendered images. Instead of three feature planes to be rasterized~\cite{chan2022efficient}, we adopt six feature planes, namely hex-planes, to model the $360^\circ$ appearance of the head from front, back, left, right, top, and bottom. Furthermore, a two-branch network is introduced to model a bald head and hair separately, achieving free-swapping of hairstyles and better fitting performance.
    
\textit{$\bullet$ Expression code $\epsilon$.}
We adopted a predefined 52-dimensional blendshapes parameter to represent facial expressions. Specifically, when establishing the \datasetname dataset, 52 expressions for each identity have been created. Therefore, the generated bilinear model inherently parameterizes the expression following the pre-defined blend shape parameters.

The overall architecture of the volume-based model is illustrated in Fig.~\ref{fig: pipeline}-(b). As with the volume-based model, we introduce RefineNet with the same setting to further improve the rendering details of the hybrid-based model.

\subsection{Point-based Model}
\label{sec: model_pt}

Very recently, 3D Gaussian splatting~\cite{kerbl20233d} has inspired point-based generative methods, which are faster in rendering and maintaining photo-realistic performance. Point-based models are more accessible to be deformed or driven than volume-based models, benefiting from their explicit structure of point clouds. The point-based 3D generative head model is formulated as:

\begin{equation}
    \left\{ \mu ,\mathbf{s},\mathbf{q},o,c \right\} = \mathbb{S}_p(p, \mathcal{G}_a(\alpha)), \textup{where\ } p \in B_p(\beta, \epsilon),
\end{equation}

\noindent where $B_p$ is a point-based bilinear model generated from the dataset. $\mathcal{G}_a$ is a 2D generative model synthesizing attribute maps from a learnable code.  $\mathbb{S}_p$ is a sampler that extracts and interpolates Gaussian parameters $\left\{ \mu ,\mathbf{s},\mathbf{q},o,c \right\}$ given the points $p$ from $B_p$.  
The training strategies of $\beta$, $\alpha$, and $\epsilon$ in the \textbf{point-based model} are explained as:

\textit{$\bullet$ Expression code $\epsilon$ and shape code $\beta$.} Similar to the volume-based model, a parameterized bilinear model is introduced to transform $\epsilon$ and $\beta$ into a 3D head shape. The difference is that the bilinear model adopted by the point-based model is represented by 3D point clouds, not 3D polygen mesh models. Both bilinear models can be animated by a predefined blendshapes parameter stream (equivalent to $\epsilon$ stream), and the head shapes are parameterized to $\beta$.  We follow the method of FaceScape~\cite{zhu2023facescape, yang2020facescape} to train the bilinear model, which is pre-trained and parameter-fixed when other parameters of the point-based generative model are trained.

\textit{$\bullet$ Appearance code $\alpha$.} Attribute maps are introduced to synthesize head appearance and are generated by a 2D generative model conditioned on $\alpha$. After an attribute map is generated, Gaussian parameters for each point generated based on $\epsilon$ and $\beta$ are extracted, rendering the final image via point-based rendering.

The overall architecture of the volume-based model is illustrated in Fig.~\ref{fig: pipeline}-(c). Although the point-based model can directly render high-resolution images due to the high rendering efficiency of 3DGS, the appearance details of the renderings are missing. So we introduce StyleUNet~\cite{wang2023styleavatar} via end-to-end training to synthesize intricate details based on the raw rendering results. Following the volume-based model, RefineNet with the same setting is also introduced further to improve the rendering details of the point-based model.  This strategy enhances the representation of fine-grained textures, such as hair and eyes, thereby improving the overall generation quality.

\begin{table*}[]
\centering
\caption{Specific designs under different representations. 
}
\begin{tabular}{@{}ccccc@{}}
\toprule
Model                & Representation                 & Shape                 & Appearance                         & Expression                       \\ \midrule
Volume-based         & NeRF          & PCA-factors           & Texture feature encoded by CNN     & Learnable code                \\
Hybrid-based         & Hex-Planes NeRF  & Morpable blend meshes & Neural texture encoded by GAN & Morpable blend meshes        \\
Point-based          &  3DGS         & Morpable blend points & Neural texture encoded by GAN & Morpable blend points        \\ \bottomrule
\end{tabular}
\label{tab:models}
\end{table*}


\section{Implementation}
\label{sec:imp}

This section introduces the methods to put the above theoretical models into practice, including training data sets (Sec.~\ref{sec: data}), training methods (Sec.~\ref{sec: train}), and how to implement image-based fitting (Sec.~\ref{sec: fit}), expression animating (Sec.~\ref{sec: ani}), and text-based editing (Sec.~\ref{sec: edit}) on the basis of the learned models. Detailed architectures for each model are provided in the supplementary material.

\subsection{Training Data}
\label{sec: data}

The \datasetname dataset encompasses $374,400$ calibrated high-resolution images rendered from $5,200$ mesh models for each identity under $52$ expressions. The ratio of males to females in the dataset is 1:1, and the age is evenly distributed between 16 and 70. The 3D heads are rendered by $72$ head-centric virtual cameras covering $3$ pitch angles and $24$ horizontal rotation angles. Both rendered images and 3D mesh models are released. Regarding model quality, the models of \datasetname dataset are more detailed and realistic than that of the Rodin dataset~\cite{wang2023rodin}, as shown in Fig.~\ref{fig: datasets}. The 3D models exhibit finer skin detail than Rodin, including pores, subtle textures, and natural variations of skin irregularities. The rendering of light and shadow on facial features is more sophisticated, capturing a complex lighting environment that includes specular reflections off the skin's surface and soft shadow edges. 

\subsection{Training Objectives}
\label{sec: train}


After introducing the \datasetname dataset, we show how the model training is implemented.  Basically, the loss functions of volume-based, hybrid-based, and point-based models comprise photometric loss and optional regularization terms. Still, the specific loss function and training methods are different.

\subsubsection{volume-based}
The loss function to train the volume-based model is formulated as follows: 

\begin{equation}
    \begin{split}
    \mathcal{L} _{volume}=\left\| \bar{c}\left( \mathbf{r} \right) -c\left( \mathbf{r} \right) \right\| _2+\left\| \hat{c}\left( \mathbf{r} \right) -c\left( \mathbf{r} \right) \right\| _2,
    \end{split}
 \end{equation}
 
\noindent where $c\left( \mathbf{r} \right), c\left( \mathbf{r} \right) $ and $\hat{c}\left( \mathbf{r} \right) $ are the ground truth, coarse predicted, and fine predicted RGB colors for ray $\mathbf{r}$ respectively, following the training strategy of vanilla NeRF~\cite{mildenhall2020nerf}. Adam optimizer~\cite{kingma2014adam} trains the model with the base learning rate of $5e-4$, decayed exponentially to $2e-5$, and the batch size of 1024 rays. The model is trained for $48$ hours on $4$ NVIDIA GTX 3090 GPUs. It takes about 10 seconds to render an image on a single 3090 GPU.

\subsubsection{hybrid-based}
Training hybrid-based model enrolls more objective terms, specifically $L1$ loss $\mathcal{L}_{L1}$, dual GAN loss~\cite{sun2023next3d} $\mathcal{L}_{dGAN}$, and density regularization $\mathcal{L}_{density}$.

$\mathcal{L}_{L1}$ loss is used to minimize the gap between the rendered image $\mathbf{I}$ and the real image $\mathbf{I}_{gt}$ , which is formulated as:

\begin{equation}
    \begin{split}
        \mathcal{L} _{L1}=\left\| \mathbf{I}-\mathbf{I}_{gt} \right\| _1,
    \end{split}
\end{equation}

The dual GAN loss $\mathcal{L}_{dGAN}$ is used to refine the detail of the synthesized image, which consists of a non-saturated adversarial loss $\mathcal{L} _G$ and a dual discriminative loss $\mathcal{L} _D$:
\begin{align}
        \mathcal{L} _G=&\mathbb{E} _{\alpha ,\beta ,\epsilon \sim \mathcal{P}}\log \left[ 1-D_d\left( G\left( \alpha ,\beta ,\epsilon \right) \right) \right]  \\
    \mathcal{L} _D=&-\left[ \begin{array}{c}
    	\mathbb{E} _{\mathbf{I}\sim \mathcal{I}}\log \left[ D_d\left( \mathbf{I} \right) \right] +
    	\frac{\gamma}{2}\mathbb{E} _{\mathbf{I}\sim \mathcal{I}}\left\| \nabla D_d\left( \mathbf{I} \right) \right\| _2\\
    \end{array} \right],
\end{align}

\noindent where appearance code $\alpha$, shape code $\beta$, and expression code $\epsilon$ are sampled from aforementioned parameter space $\mathcal{P}$, ground truth images are sampled from $\mathcal{I}$, $G(\cdot)$ denotes the generator in hybrid-based model, and $D_d(\cdot)$ is dual-discrimination scheme in EG3D\cite{egger20203d}. $\nabla D_d (\cdot)$ is R1 regularization in GAN training and $\gamma$ is the hyperparameter. Therefore, the total $\mathcal{L}_{dGAN}$ is formulated as:
\begin{equation}
    \begin{split}
        \min_G \max_D \mathcal{L} _{dGAN}\left( D,G \right) =\mathcal{L} _G+\mathcal{L} _D,
    \end{split}
\end{equation}
Finally, density regularization utilizes the total variation loss $\mathrm{TV}(\cdot)$ to the density $\sigma$ to ensure the smooth and realistic geometry in the radiation field:
\begin{equation}
    \begin{split}
        \mathcal{L} _{density}=\mathrm{TV}\left( \sigma \left( x \right) ,\sigma \left( x+\Delta x \right) \right) ,
    \end{split}
\end{equation}
\noindent where $x$ is coordinates sampled in hex-planes.

Overall, the training objective for hybrid-based model is formulated as follows:
\begin{equation}
    \begin{split}
        \mathcal{L} _{hybrid}=\mathcal{L} _{L1}+\lambda _1\mathcal{L} _{dGAN}+\lambda _2\mathcal{L} _{density},
    \end{split}
\end{equation}

\noindent $\lambda_{1}$ is set to $0.01$. The density regularization weight $\lambda_{2}$ is activated iteratively, following the setting of EG3D. Adam optimizer is used to train the model with the base learning rate $l_{base}$ as $0.0025$ and the batch size as $8$. The model is trained for 10 days on 4 NVIDIA GTX 3090 GPUs and generates head renderings at several minutes to get shape mesh and neural texture, rendering at 15 FPS.

\subsubsection{point-based}
The point-based model is first generated Gaussian attribute maps $\mathcal{M}$, and then render a coarse result via 3DGS, denoted $\mathbf{I}_{raw}$. We denote the result after $\mathbf{I}_{raw}$ is processed by subsequent UNet as $\mathbf{I}$. In the point-based approach, the fitting loss supervises both $\mathbf{I}_{raw}$ and $\mathbf{I}$. In a slight abuse of notation, we reuse $\mathcal{L}_{L1}$:
\begin{align}
\mathcal{L} _{L1}&=\left\| \mathbf{I}_{raw}-\mathbf{I}_{gt} \right\| _1+\left\| \mathbf{I}-\mathbf{I}_{gt} \right\| _1,
\end{align}

We similarly apply GAN loss $\mathcal{L} _{GAN}$ to enhance the details. Since $I_{raw}$ and $I_{gt}$ have the same size in the point-based model, we do not need to use dual-discrimination scheme.

Following GGHead\cite{kirschstein2024gghead}, we stabilize the training along two regularization terms on Gaussian attributes. We use L2 regularization for the Gaussian scale map $\mathcal{M}_{s}$ and the Gaussian offset map $\mathcal{M}_{o}$ to keep the Gaussians within a reasonable range during training:
\begin{equation}
    \begin{split}
    \mathcal{L} _{scale}=\left\| \mathcal{M} _{s} \right\| _2,
    \\
    \mathcal{L} _{offset}=\left\| \mathcal{M} _{o} \right\| _2,
    \end{split}
\end{equation}

Overall, the loss function to train the point-based model is formulated as follows:

\begin{align}
\mathcal{L} _{point}=\mathcal{L} _{L1}+\lambda _1\mathcal{L} _{GAN}+\lambda _2\mathcal{L} _{scale}+\lambda _3\mathcal{L} _{offset},
\end{align}

\noindent where $\lambda_1,\lambda_2,\lambda_3$ are set as 1.0, 0.1, 1.0 respectively. The model is trained for around 2 days on 8 NVIDIA A6000 GPUs and generates head renderings at 30 FPS in NVIDIA RTX 3090.


\subsection{Single-Image Fitting}
\label{sec: fit}

\begin{figure}[tb]
    \centering
    \includegraphics[width=1\linewidth]{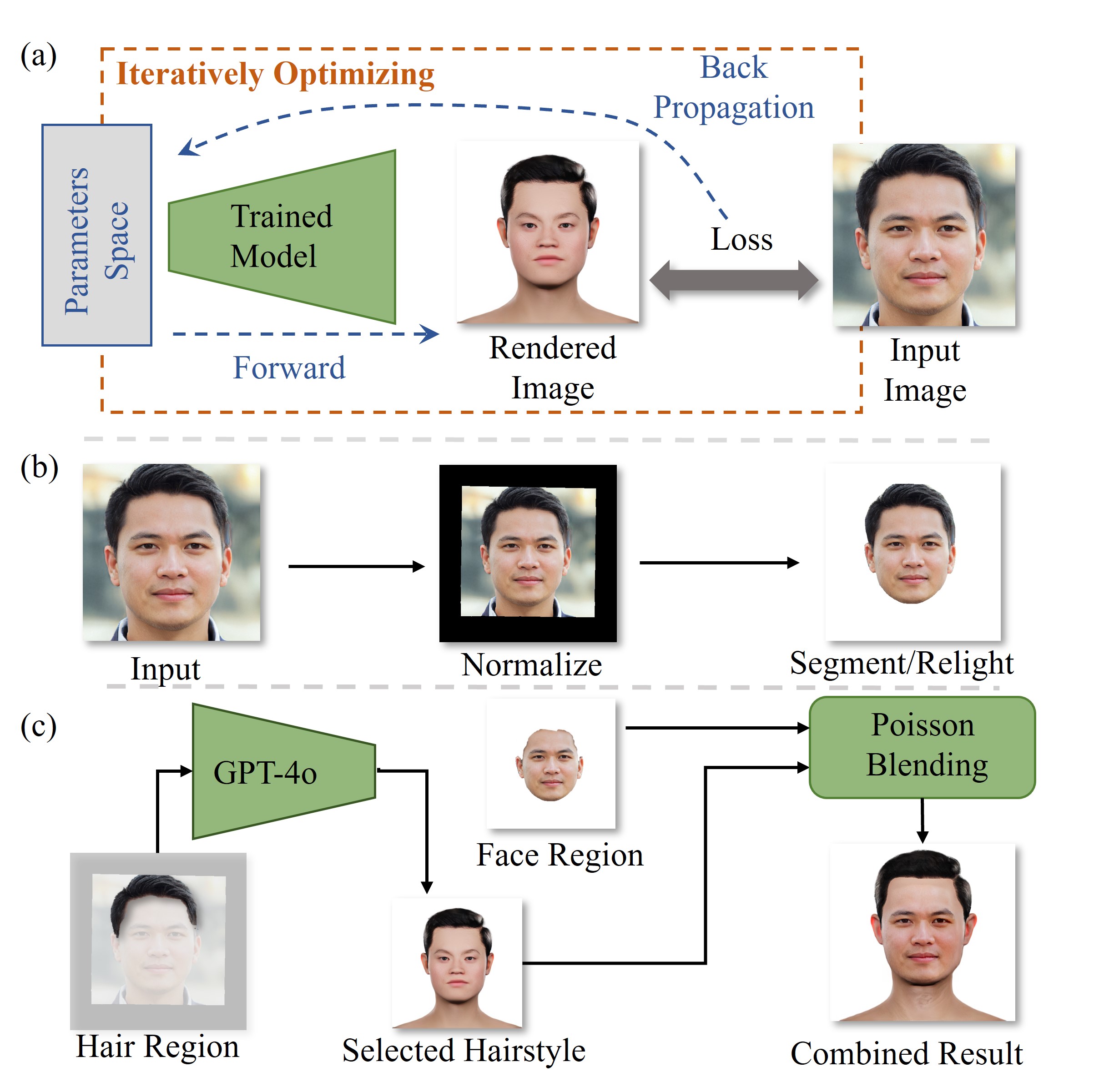}
    \caption{
    \textbf{Visualization of our fitting pipeline.}  (a) provides an overview of our fitting pipeline, which iteratively optimizes values in the parameters space to fit an in-the-wild image. After optimizing, we create a 3D head model in our model space, enabling rigging and further editing applications.\
    (b) illustrates the pre-processing step to normalize the input image, bringing it closer to the synthetic dataset distribution.\
    (c) presents our approach to fitting the full head, separately considering the hair and face regions. We use GPT-4o to select the closest hairstyles in our \datasetname and combine it with a random initialized appearance code, as well as initialize the face shape with Poisson blending. This creates a new reference image for full-head optimization. \
    }
    \label{fig: pipe_fit}
\end{figure}

Single-image fitting involves finding a latent code in the trained generative model's parametric or feature space most similar to a target image, as illustrated in Fig.~\ref{fig: pipe_fit}.  Such a latent code can be a certain parametric code, an optimized neural texture map, or an attribute map.   The fitting results support free-view rendering, text-based editing, and expression rigging.  The fitting methods of the generative models under different representations are different and customized to achieve the best fitting performance.  

To fit the \textit{volume-based} model, we simply optimize $\beta, \alpha, \epsilon$ through backward propagation, keeping the network parameters fixed. 
For better fitting results, we preprocess the input image by normalizing the face position, segmenting the head region using ~\cite{luo2020ehanet}, and removing the uneven lighting with ~\cite{zhou2019deep}, as shown in Fig.~\ref{fig: pipe_fit}-(b).
$\beta$ and $\alpha$ are initialized randomly using a Gaussian distribution, while $\epsilon$ is initialized with the learned latent of neutral expression. We then freeze the pre-trained network weights and optimize $\beta, \alpha, \epsilon$  through the network by minimizing the MSE loss function between the predicted and target colors of sampled rays.

The fitting method for the \textit{hybrid-based} models includes additional preprocessing, as illustrated in Fig.~\ref{fig: pipe_fit}-(c).
Considering that the hairstyles in the \datasetname dataset are quite limited, it is impossible to fit the trained generative model to the diverse in-the-wild hairstyles. So, we compromise the hairstyle fitting, leveraging GPT-4o to predict a closest hairstyle to the target image from the synthetic dataset.  Then, the fitted hairstyle and the segmented face are merged with Poisson blending~\cite{perez2023poisson} to create a synthesized target image for fitting head appearance. The shape code $\beta$ is optimized by minimizing the error distance between the parametric face and the estimated 3D facial landmarks, following the fitting method of FaceScape~\cite{zhu2023facescape, yang2020facescape}. We then optimize the neural texture via GAN-inversion by minimizing the difference between the synthesized target image and the generated image. 
Compared to optimizing appearance code $\alpha$, optimizing the neural texture enables fitting in a much larger solution space, thus enhancing the fitting performance.

The fitting method of the \textit{point-based} method inherits most operations from the hybrid-based method with some differences. We observed that directly optimizing Gaussian attribute maps often leads to over-fitting to the target image, resulting in irregular 3D geometry~\cite{zhang2024cor}. To avoid this inherent nature of 3D Gaussian splats, we optimize the style vector $w$ instead of appearance code $\alpha$ for better fitting performance.

When fitting to an in-the-wild image, several disparities exist between it and the synthetic data, including the uncertain facial region position, unnecessary background, and complex lighting conditions. A pre-processing pipeline is introduced to bridge these gaps. Firstly, we normalize the face region by extracting the 2D landmarks $L_t$ from the target image~\cite{kazemi2014one} and align $L_t$ to the predefined 3D landmarks $L_c$ of the canonical face using an affine transformation~\cite{yang2020facescape}. Subsequently, the background is removed using EHANet~\cite{luo2020ehanet, CelebAMask-HQ}, and the lighting is normalized with the relighting method~\cite{zhou2019deep}. This pre-processing stage is crucial for mitigating lighting influence and ensuring a consistent starting point for all the above fitting methods.

\begin{figure*}[tb]
    \centering
    \includegraphics[width=1.0\linewidth]{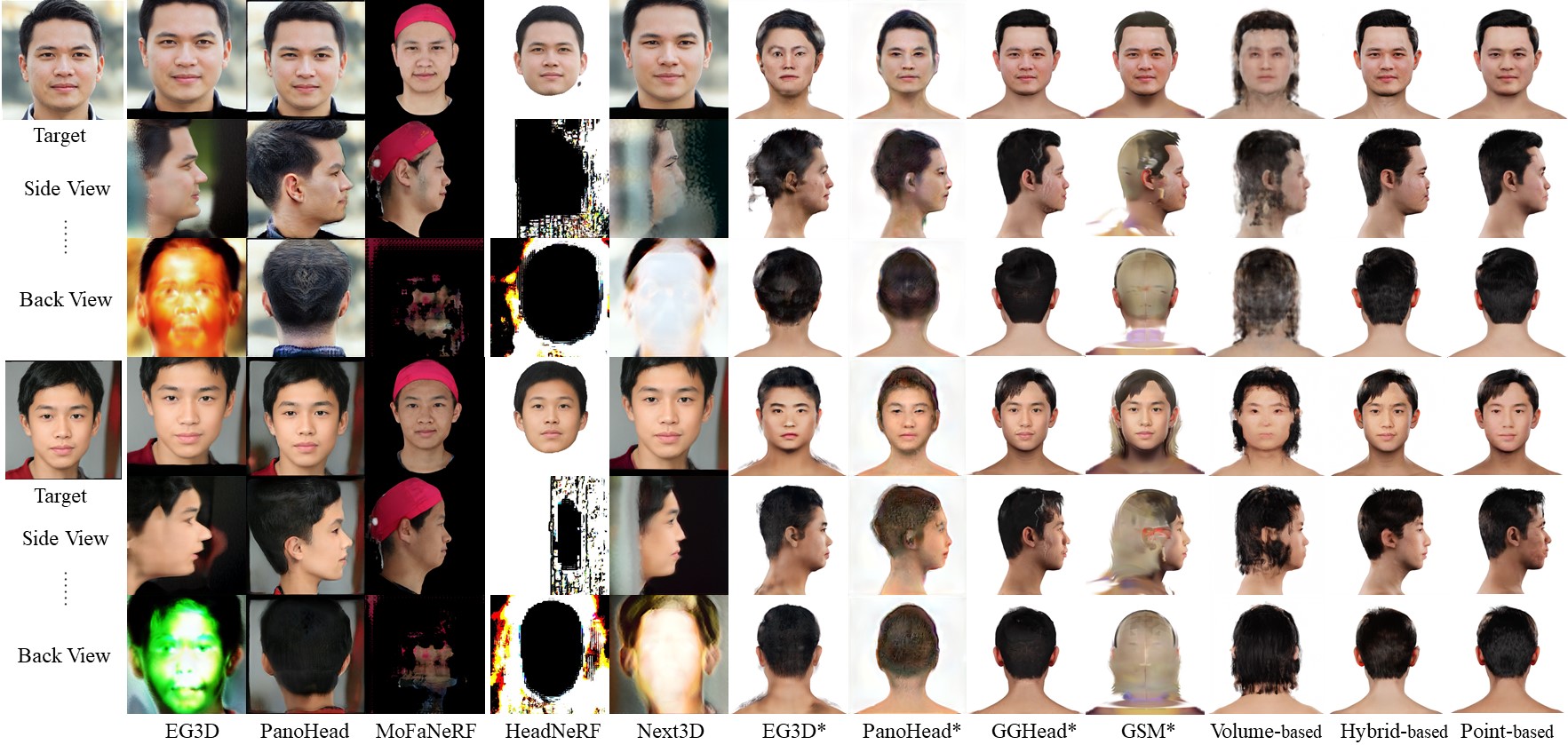}
    \caption{
    \textbf{Comparison of fitting results on in-the-wild images.} We compare our method with previous parametric or generative 3D head models in single-image fitting. For a comprehensive comparison, both original models and re-trained models are compared. 
    }
    \label{fig: comp_fit}
\end{figure*}

\begin{figure*}[tb]
    \centering
    \includegraphics[width=0.75\linewidth]{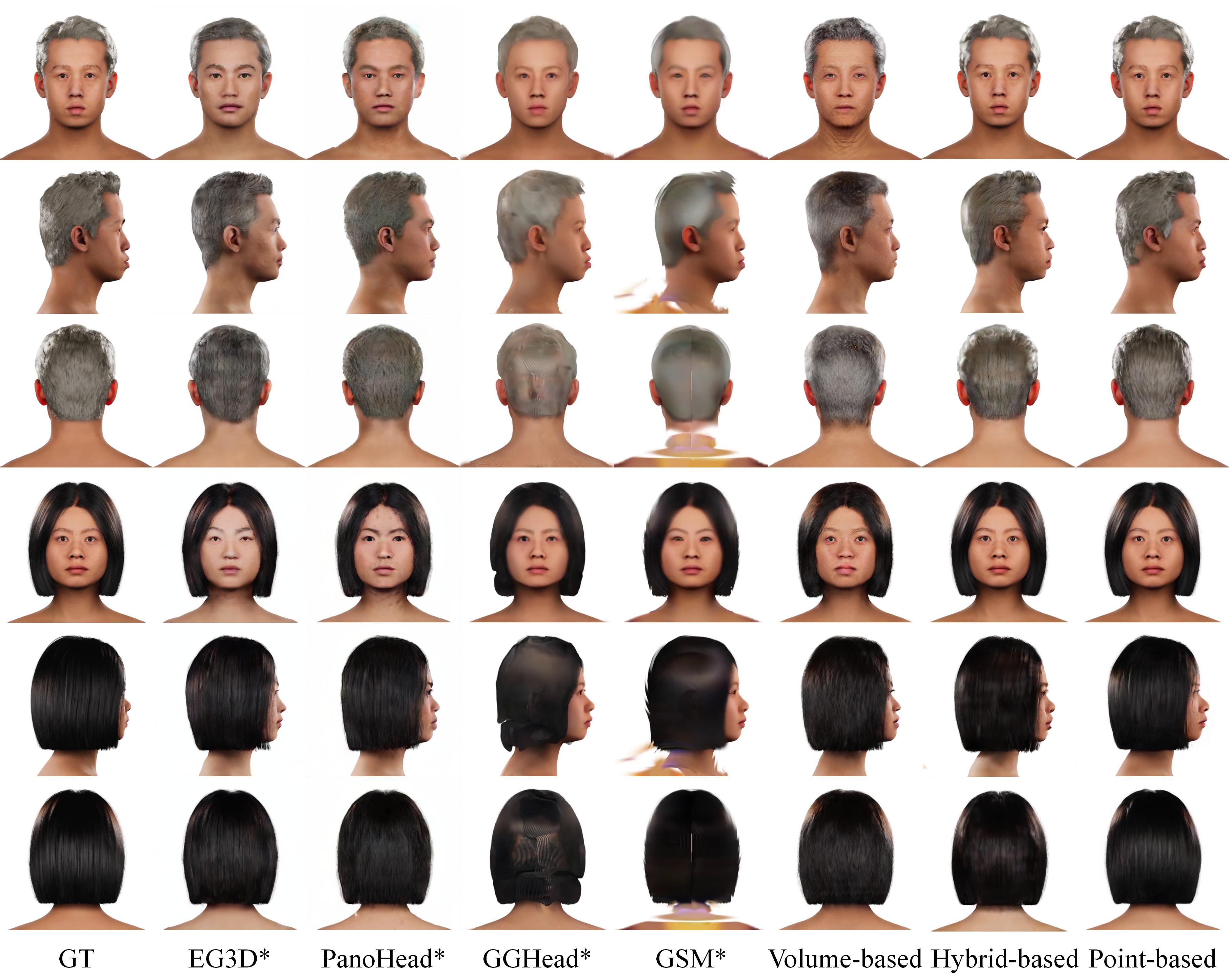}
    \caption{
    \textbf{Comparison of fitting results on \datasetname.} We compare our method with previous parametric or generative 3D head models in single-image fitting. For a comprehensive comparison, we retrain the other baselines on the proposed dataset. 
    }
    \label{fig: comp_fit_syn}
\end{figure*}

\begin{table}[tb]
\centering
\caption{\textbf{Quantitative evaluation of fitting results in synthetic dataset.
}
}
\begin{tabular}{@{}lcccc@{}}
\toprule
Method     & PSNR(dB)↑  & SSIM↑($\times0.1$)       & LPIPS↓($\times0.1$)     \\ \midrule
EG3D*~\cite{chan2022efficient}  &19.81±2.60 &7.67±0.66 &1.50±0.75\\
PanoHead*~\cite{an2023panohead}  &20.00±2.76 &7.71±0.59  &1.51±0.71 \\
GGHead*~\cite{kirschstein2024gghead}  & 17.49±3.1&7.74±0.42  & 3.14±0.51 \\
GSM*~\cite{abdal2024gaussian}  & 19.59±0.44&7.79±0.36  & 3.34±0.40 \\
\midrule
Volume-based  &20.58±2.84  &7.64±0.61   &1.43±0.54\\
Hybrid-based & 20.87±1.8 & 7.83±0.46 & 1.52±0.43 \\
Point-based & \textbf{23.19±1.5} & \textbf{7.97±0.56} & \textbf{1.23±0.34} &
\\\bottomrule
\end{tabular}
\label{tab: fit}
\end{table}

\begin{table}[t]
\centering
\caption{\textbf{Quantitative evaluation of generated results.} }
\begin{tabular}{@{}lccc@{}}
\toprule
Method     & PSNR(dB)↑  & SSIM↑($\times0.1$)       & LPIPS↓($\times0.01$)        \\ \midrule
EG3D*~\cite{chan2022efficient}      & 26.45±3.27 & 8.50±0.45 & 6.60±2.24 \\
PanoHead*~\cite{an2023panohead}  & 25.89±3.88 & 8.60±0.46 & 10.26±2.83 \\
GGHead*~\cite{kirschstein2024gghead}&  26.03±1.97 &  8.75±0.40 &  16.12±4.28\\
GSM*~\cite{abdal2024gaussian}&  21.14±2.11 &  8.45±0.38 &  22.10±5.09\\
Volume-based & 28.09±2.36 & 8.60±0.32 & 7.45±1.73 \\
Hybrid-based & 28.27±2.61 & 8.66±1.42 & 6.31±1.71 \\
Point-based & \textbf{30.00±1.80} & \textbf{8.75±0.36} & \textbf{3.56±1.20} \\

\bottomrule
\end{tabular}
\label{tab: comp}
\end{table}

\begin{figure*}[t]
    \centering
    \includegraphics[width=1.0\linewidth]{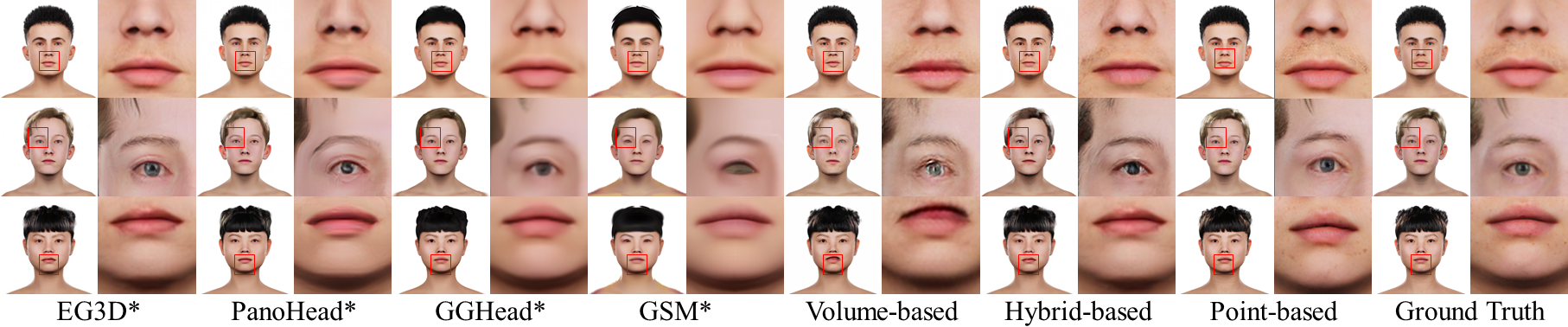}
    \caption{\textbf{Visual comparison of generated results.} We compare our method with other retrained models to demonstrate our representation ability. 
    }
    \label{fig: comp_gen}
\end{figure*}

\begin{table}[t]
\centering
\caption{\textbf{Quantitative evaluation of generated results for the ablation study.} SSIM is scaled by a factor of $0.1$, while LPIPS is scaled by a factor of $0.01$.}
\begin{tabular}{clccc}
\toprule
Method & Experiment & PSNR(dB)↑ & SSIM↑ & LPIPS↓ \\ \midrule
\multirow{3}{*}{Volume-based} 
 & w/o ISM & 27.48±1.51 & 8.56±0.38 & 8.11±2.00 \\
 & w/o Refine & \textbf{29.78±1.32} & \textbf{8.96±0.30} & 16.73±4.76 \\
 & Full & 28.06±1.58 & 8.61±0.38 & \textbf{7.78±1.94} \\ \midrule

 \multirow{4}{*}{Hybrid-based} 
 & Vani-NeRF & 24.44±2.65 & 8.26±0.37 & 8.16±1.68 \\
 & Tri-planes & 25.66±2.14 & 8.27±0.13 & 9.54±1.72 \\
 & w/o Refine & \textbf{28.92±2.32} & \textbf{8.83±0.13} & 7.31±1.75 \\
 & Full & 28.27±2.61 & 8.66±1.42 & \textbf{6.31±1.71} \\ \midrule

  \multirow{3}{*}{Point-based} 
 & w/o U-Net & 26.97±2.17 & \textbf{8.89±0.36} & 16.23±5.66 \\
 & w/o Refine & 27.31±1.93 & 8.58±0.45 & 5.34±1.78 \\
 & Full & \textbf{30.00±1.80} & 8.75±0.36 & \textbf{3.56±1.20} \\ 
 \bottomrule
\end{tabular}
\label{tab: ablation}
\end{table}

\begin{figure*}[th]
    \centering
    \includegraphics[width=1.0\linewidth]{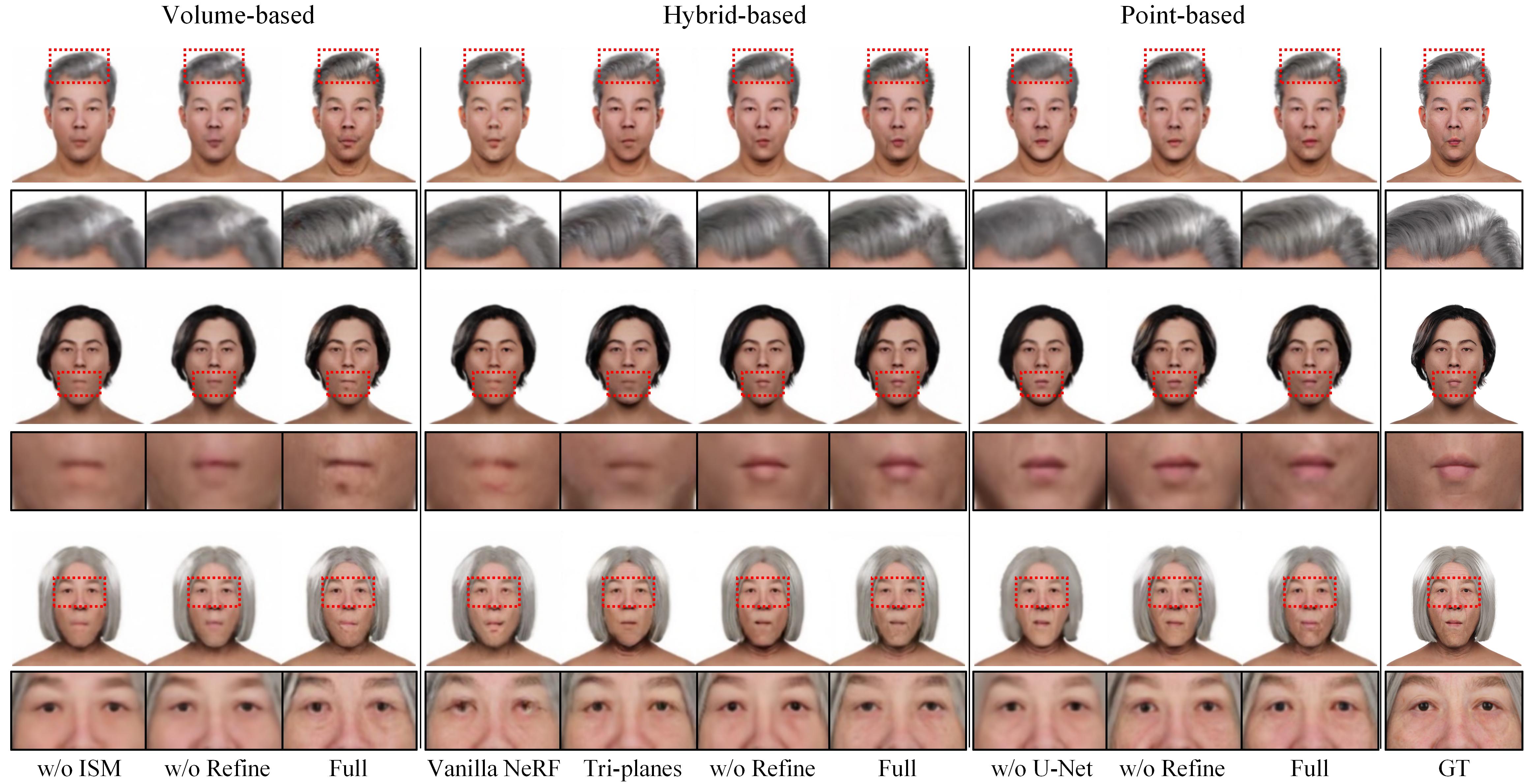}
    \caption{\textbf{Ablation study.} 
    The performance of our full model outperforms all the ablated settings with more clear and detailed rendering, which proves the effectiveness of these proposed modules. The issues are highlighted with the red dotted box. 
    }
    \label{fig: ablation}
\end{figure*}

\begin{figure*}[th]
    \centering
    \includegraphics[width=1.0\linewidth]{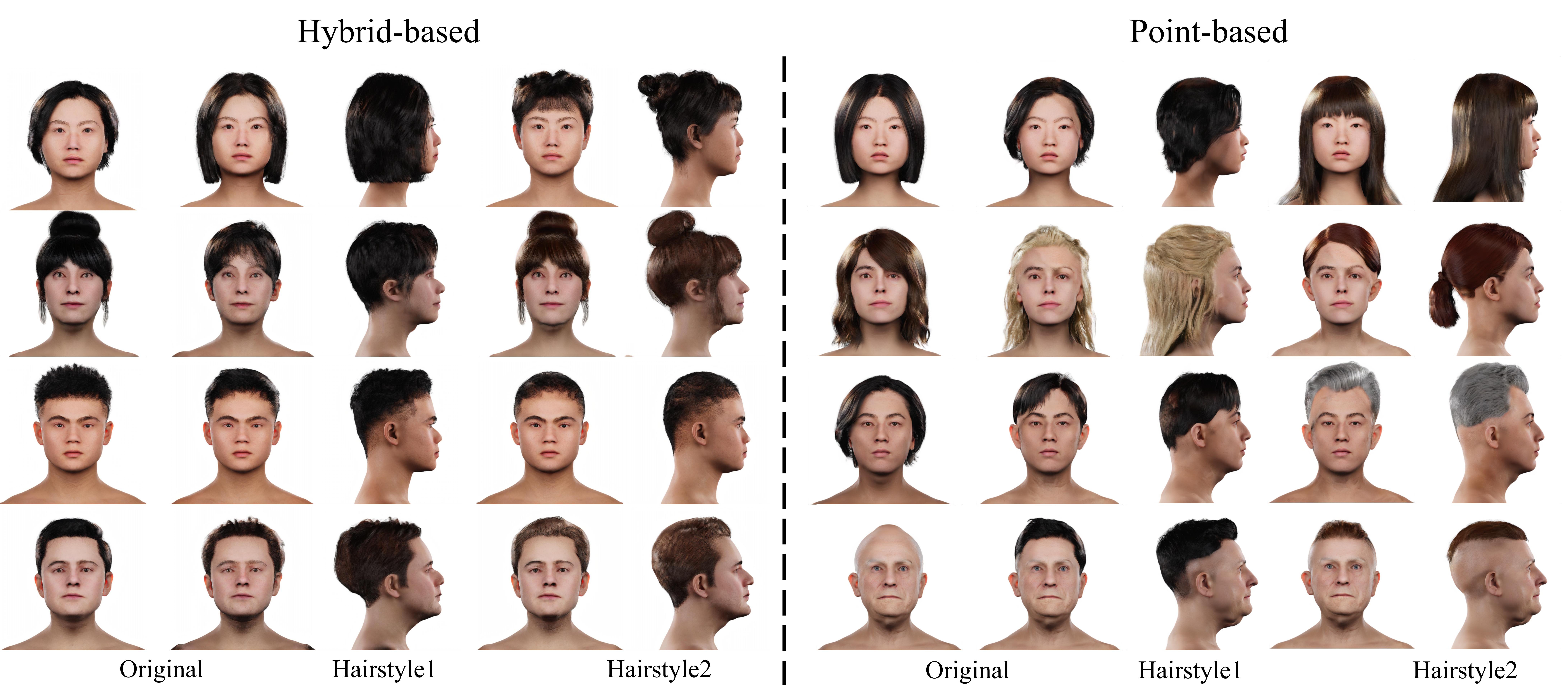}
    \caption{\textbf{Hair swapping results.} Given a generated head, the original hairstyle can be replaced with other hairstyles.  
    }
    \label{fig: hair_and_anim}
\end{figure*}

\begin{figure*}[th]
    \centering
    \includegraphics[width=1.0\linewidth]
    {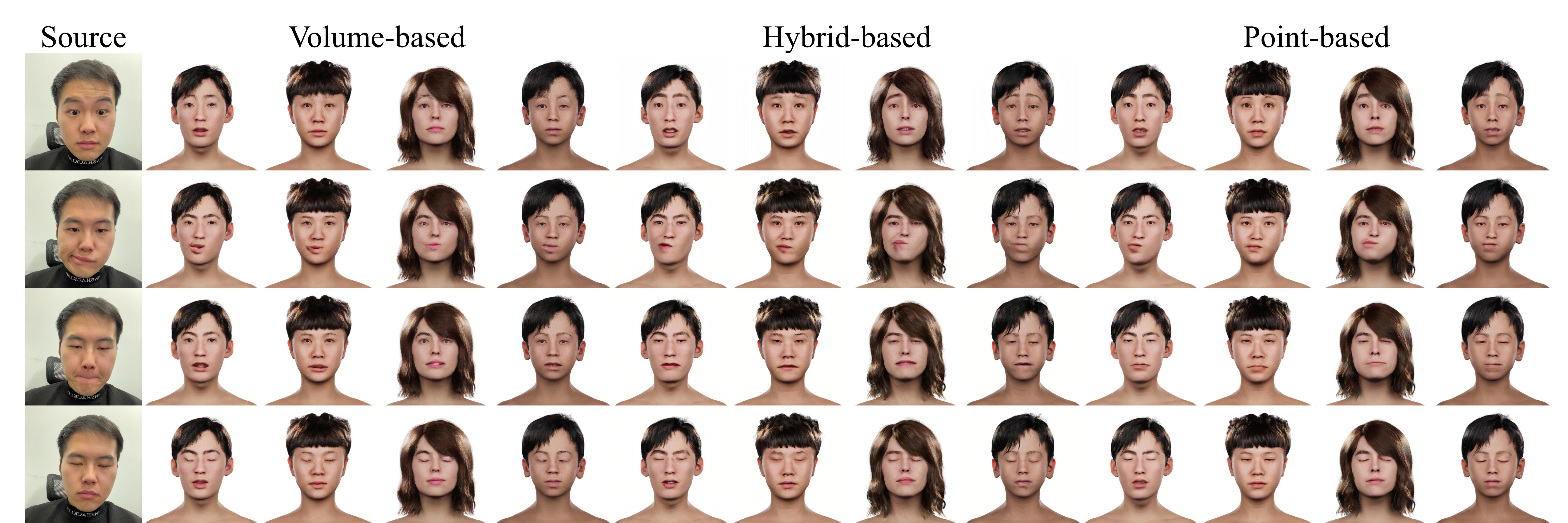}
    \caption{\textbf{Animated results.} Left: Given a generated head, the original hairstyle can be replaced with other hairstyles. Right: our generated head can be driven by blendshapes parameters, with facial details accurately synthesized.
    }
    \label{fig: hair_and_anim}
\end{figure*}

\begin{figure*}[th]
    \centering
    \includegraphics[width=0.8\linewidth]{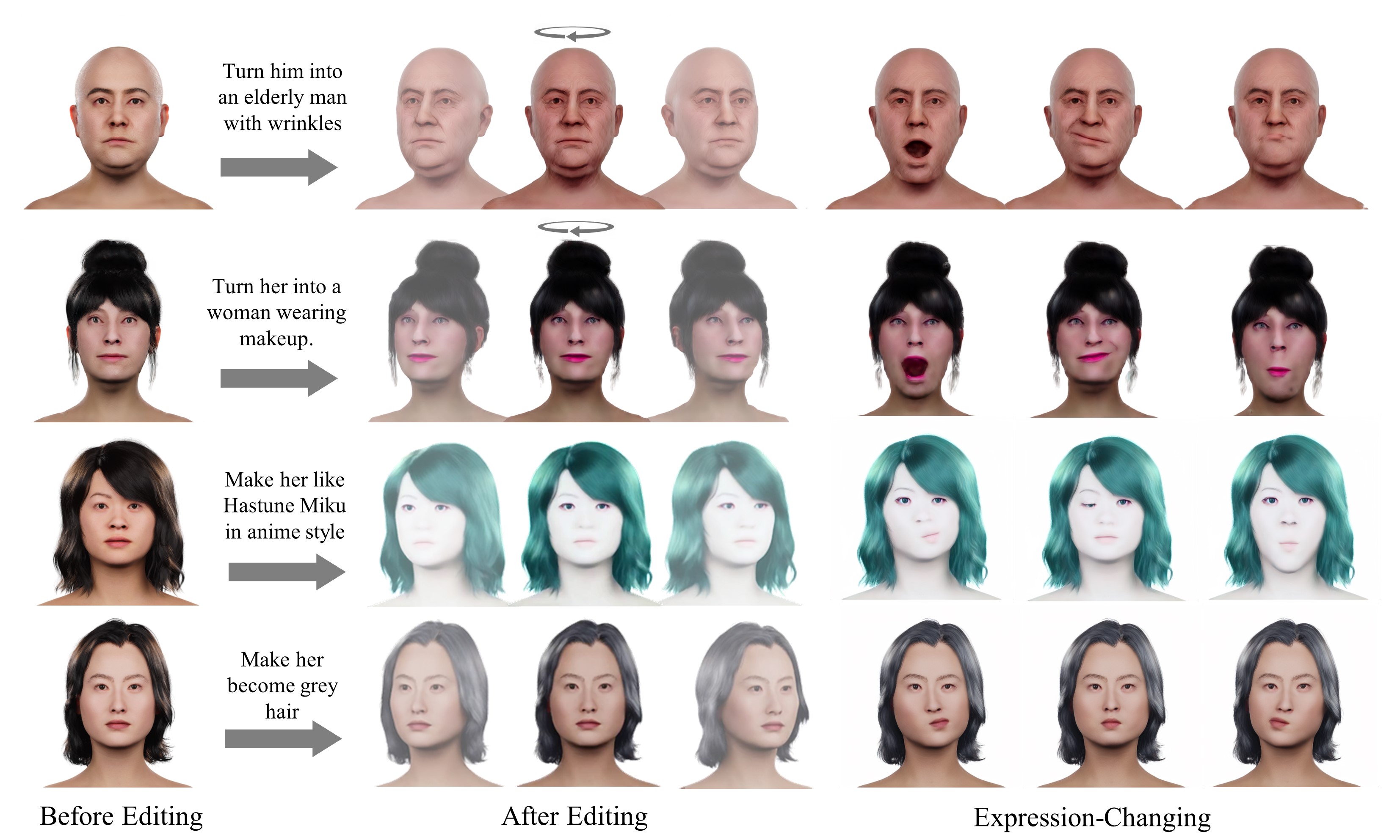}
    \caption{\textbf{Text-based editing results.} Given a text-based input (italics sentences), our generated model can be further edited to match the prompt while keeping identities unchanged. The animated frames of the edited head are shown on the right.
    }
    \label{fig: text}
\end{figure*}

\subsection{Animation}
\label{sec: ani}

The generated or fitted head by our models can be directly animated since our parametric 3D mesh is defined following a standard $52$ facial blendshapes, which is also adopted by ARKit~\cite{arkit}. Specifically, by putting the blendshape parameter stream to the generative model as $\beta$ and keeping $\beta$ and $\alpha$ fixed, an identity-consistent 3D head animation can be generated.
Our model can synthesize a temporally consistent appearance not closely attached to the 3D surface, such as hair, eyelashes, and the inner mouth. At the same time, our animated model can synthesize dynamic details, such as smiling dimples and brow-raising wrinkles, which are hard to represent by textured mesh models.  These features are not available in mesh-based 3D head morphable models.

\subsection{Text-based Editing}
\label{sec: edit}

We propose a text-based fitting approach for hybrid-based and point-based generative models.  
Inspired by Instruct-NeRF2NeRF~\cite{instructnerf2023}, we incorporate text-based image-conditioned diffusion model~\cite{Brooks_2023_CVPR} with our animatable 3D parametric representation for 3D head editing.  
Specifically, after a head is generated or fitted, the multi-view images are rendered and edited by a text-guided image-to-image translator. The edited images are then used to fine-tune our parametric 3D head model. Empirically, we sample patches on generated images to fine-tune our parametric model, leading to better performance than the original random sampling strategy. Our text-based editing strategy aims to add abstract features, such as `wearing makeup' and `aged', to a generated head, without changing his/her identity.  The edited head can still be animated by blendshape parameters.

The above method works for the hybrid-based model but not the point-based model. The reason is that we cannot mix signals from edited and unedited images, as 3DGS does not use rays for training. Instead, it requires rendering full images. Therefore, we directly edit the multi-view images of chosen identity using Instruct-Pix2Pix based on given prompts. In our experiments, we found that editing images from horizontal view orbits yields more stable results. Afterward, we finetune StyleUNet in the point-based model using the edited multi-view images. This essentially performs a style transfer through the StyleUNet, and experiments yield that this approach exhibits 3D-consistent editing.

\section{Experiments}
\label{sec: exp}

In this section, we first compare our fitting results and generated results with previous methods, then analyze the effectiveness of the proposed models through the ablation study, and finally show the results of animation and text-based editing.
Peak Signal-to-Noise Ratio (PSNR)~\cite{hore2010image}, Structural Similarity Index (SSIM)~\cite{wang2004image}, and Learned Perceptual Image Patch Similarity (LPIPS)~\cite{zhang2018unreasonable} are used for quantitatively evaluating the image quality.

\subsection{Comparison of single-image fitting}
We compare our methods with state-of-the-art generative 3D head models in the task of single-image fitting, including EG3D~\cite{chan2022efficient}, PanoHead~\cite{an2023panohead}, HeadNeRF~\cite{hong2022headnerf},  Next3D~\cite{sun2023next3d}, GSM~\cite{abdal2024gaussian}, and GGHead~\cite{kirschstein2024gghead}. 
To fairly compare the performance and effectiveness, we retrain these methods on the \datasetname dataset and denote these baselines with a symbol `$*$'. 
Since GSM is trained on the full-body dataset and GGHead has not published their weights, we only report their results with retrained models.

For quantitative analysis, we refer to Fig.~\ref{fig: comp_fit} for in-the-wild inputs and Fig.~\ref{fig: comp_fit_syn} for the \datasetname dataset. Additionally, we perform a qualitative comparison in Tab.~\ref{tab: fit} using the \datasetname dataset. We fit the front-view image and then select 24 views (covering a 360° horizontal rotation) as the ground truth to evaluate the 3D quality of the fitted results.

EG3D, PanoHead, and Next3D are GAN-based methods trained on a large-scale dataset of 2D face images. Although these methods can reconstruct relatively great results for the given views, they fail to synthesize plausible results when viewing angles $> 90^\circ$. PanoHead produces plausible results at the back views by adding in-house captured head images from the side views, but the fitted head is not riggable. 
The original MoFaNeRF model, trained on studio-captured high-quality 3D face models, also produces unsatisfactory results when viewing angles are larger than $90^\circ$ and is limited due to the presence of headgear.  HeadNeRF leverages additional in-the-wild images to improve results at the front but severely degrades when viewing angles are larger. 
In conclusion, these models are limited to their training dataset and cannot generate a $360^\circ$ animatable head.
To eliminate the effects of the dataset and fairly compare the performance of these methods, we retrain them on the \datasetname dataset.  

For volume-based and hybrid-based models, EG3D* and PanoHead* cannot fully align with the given image because they optimize their latent space $w$ to fit the given images, which is limited by the scale of the dataset. Our volume-based method does not separate the modeling of hair and bald head, resulting in limited representation ability for in-the-wild images. By comparison, the results of the hybrid-based model are visually better, aligning well with the front view and plausible in the side and back views.

For point-based models, we found that direct rendering with 3DGS fails to produce fine results~\cite{zheng2024headgap}, which is also verified by our ablation study. The ultra-fine texture details in \datasetname dataset pose a challenge for 3DGS modeling due to the inherent low-pass characteristics of Gaussian splats. 
As a result, GSM struggles to capture the details effectively with its fixed-position hierarchical Gaussian, resulting in numerous shell-like artifacts. GGHead also faces similar problems, leading to blurry outputs. Our point-based model, enhanced with a U-Net-based neural renderer, achieves better results by synthesizing the image more effectively.

\subsection{Evaluation of Generated Results}

To evaluate the representation of the generative models, we compare the generated images with the ground-truth training images on a subset containing 100 images with different identities.  For a fair comparison, we retrain the previous models on our \datasetname dataset, including EG3D~\cite{chan2022efficient}, PanoHead~\cite{an2023panohead}, GGHead~\cite{kirschstein2024gghead} and GSM~\cite{abdal2024gaussian}. 
The quantitative and qualitative evaluations are reported in Fig.~\ref{fig: comp_gen} and Tab.~\ref{tab: comp}.  
EG3D and PanoHead are static generative head models that utilize tri-planes and tri-grids to compress volumetric features for high-speed rendering.  Qualitative and quantitative experiments show that they can't synthesize faithful details of facial appearance. 
GGHead synthesizes UV maps of Gaussian attributes and places the primitives relative to the template mesh, which struggles to reproduce high-frequency details and produces blurred results. GSM fails to represent the complex hair typology since it uses a fixed-position hierarchical shell map.
By contrast, our models outperform previous models in generating accuracy. The volume-based model still generates unclear results due to the low-pass nature of the MLPs \cite{tancik2020fourfeat}. Hybrid-based methods leverage neural texture and hex-planes to synthesize more textural details on the base of an explicit morphable mesh model. The point-based method avoids representing the whole volume space but directly synthesizes the explicit 3D head with 3D Gaussian Splats, achieving superior performance. As the hair and head are detached in the feature space, our model supports free hair-swapping for a generated or fitted head. The hair-swapping results and interpolation of shape code $\beta$ and appearance code $\alpha$ are shown in the supplementary video. 

\subsection{Ablation Study}

We study the effectiveness of the proposed modules on generating quality by ablating certain modules, as reported in Tab.~\ref{tab: ablation} and Fig.~\ref{fig: ablation}.  First, we examine the impact of the RefineNet module, which is applied to all three representations: volume-based, hybrid-based, and point-based. We ablate the RefineNet, denoting it as `w/o Refine'. Although the RefineNet slightly decreases PSNR and SSIM scores, it generates more details for better visualized results, as demonstrated with higher LPIPS scores. This occurs because the details synthesized by the RefineNet may not perfectly match the ground truth details.

The other ablation experiments are conducted for volume-based, hybrid-based, and point-based separately. For the volume-based model, we find that the removal of the ISM (`w/o ISM') causes a degradation in performance. For the hybrid-based method, we replace the neural texture and hex-planes with a conditioned vanilla NeRF, denoting it as `Vani-NeRF'. We also replace the hex-plane representation with a classic tri-plane representation, denoting it as `Triplanes'. These comparisons effectively demonstrate the efficiency of the proposed hex-planes architecture.  For the point-based model, we removehe neu tral renderer with U-Net architecture (`w/o U-Net') that decodes the Gaussian splat features into colors. 
Instead, the Gaussian splatting color is directly predicted without decoding through a U-Net. This leads to performance degradation, as optimizing the high reflection color of hair is quite challenging for Gaussian splats only.

In a comprehensive comparison of the full models of volume-based, hybrid-based, and point-based methods, our method progressively enhances the representation ability. The point-based method not only provides better visual results, including the deformation and details but also outperforms the other approaches.

\subsection{Applications}

As mentioned above, our learned models can be used for 3D animation (Section~\ref{sec: ani}) and text-based editing (Section~\ref{sec: edit}).  Fig.~\ref{fig: hair_and_anim} shows our animated results driven by a blendshape parameter stream captured by ARKit software, and the video results are shown in the supplementary video.  We can see that facial details like mouth, eyes, and motion wrinkles are synthesized vividly.  Fig.~\ref{fig: text} shows our generated 3D models can be further edited given a text prompt. The subject's identity is maintained after the editing, while abstract features like `makeup' and `aged' are synthesized.

\section{Conclusion}
\label{sec:con}

In this paper, we study to learn a native 3D generative model for 360$^\circ$-renderable head avatar from limited but high-accuracy 3D head datasets. Volume-based, hybrid-based, and point-based models are explored with appearance, shape, and motion disentangled in a parametric space. The proposed models can generate full 3D heads with hair. We further propose customized single-view fitting, text-based editing, and animation methods for downstream applications.  

\noindent \textbf{Limitations.} 
Creating high-fidelity avatars comes at a considerable cost, thus limiting our dataset to 100 identities. Considering the high cost of producing high-fidelity digital human data, we think it is meaningful to study how to learn a parametric head model with a limited data amount.  Although some in-the-wild fitting results are plausible by learning from this finite training set, stability in fitting remains a challenge. Evidence of this can be found in the failure case presented in our supplementary material.  A potential solution might entail expanding the dataset and incorporating real-world images and models into the training process.
Moreover, our method does not decompose the material and lighting, which results in some highlights being baked into the texture. This reduces the photorealism and application potential and could be alleviated by utilizing the illumination and material information from the synthetic dataset in the future. 

\section*{Acknowledgments}
This study was funded by NKRDC 2022YFF0902200, NSFC 62441204, and Tencent Rhino-Bird Research Program. We thank Tobias Kirschstein, and other authors of GGHead for the discussions and valuable suggestions.

\bibliographystyle{IEEEtran}
\bibliography{main}


\vfill

\newpage

\appendix

The supplementary material contains a video and a PDF file. The video shows the dynamic results of our methods and the comparison with other state-of-the-art works. In the PDF file, we first present the implementation details in Sec.~\ref{sec:implement}, encompassing an overview of the utilized dataset and a description of the three types modeling methods. Then, more experimental results and visual results are shown in Sec.~\ref{sec:more_exp} and Sec.~\ref{sec:more_res}, respectively. Finally, failure cases and limitations are discussed in Sec.~\ref{sec:limi}. 


\section{Implementation Details}
\label{sec:implement}


\subsection{Details of the Dataset}
\begin{figure*}[!htbp]
    \centering
    \includegraphics[width=1\linewidth]{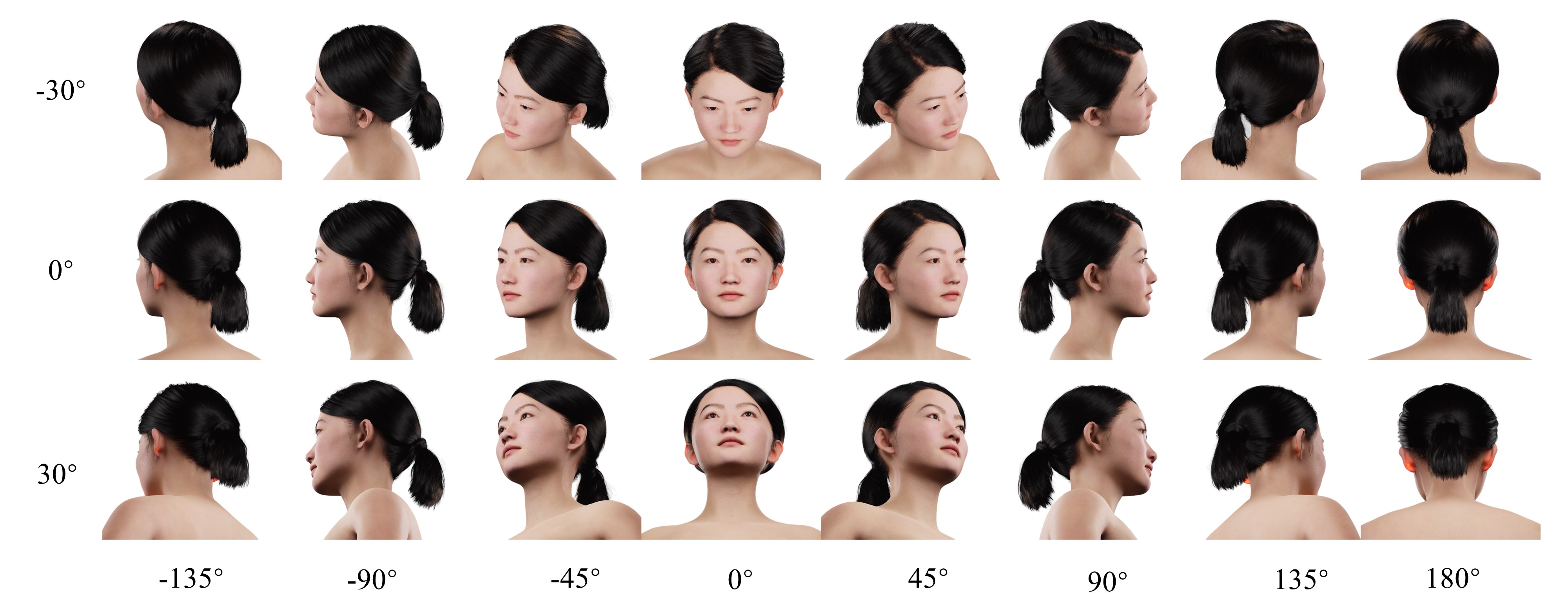}
    \vspace{-0.1in}
    \caption{
    Multi-view images are rendered at 72 views with 3 pitch angles and 24 yaw angles. Here are some of the rendering views.
    }
    \vspace{-0.1in}
    \label{fig:supp_multiview}
\end{figure*}

\begin{figure*}[!htbp]
    \centering
    \includegraphics[width=1\linewidth]{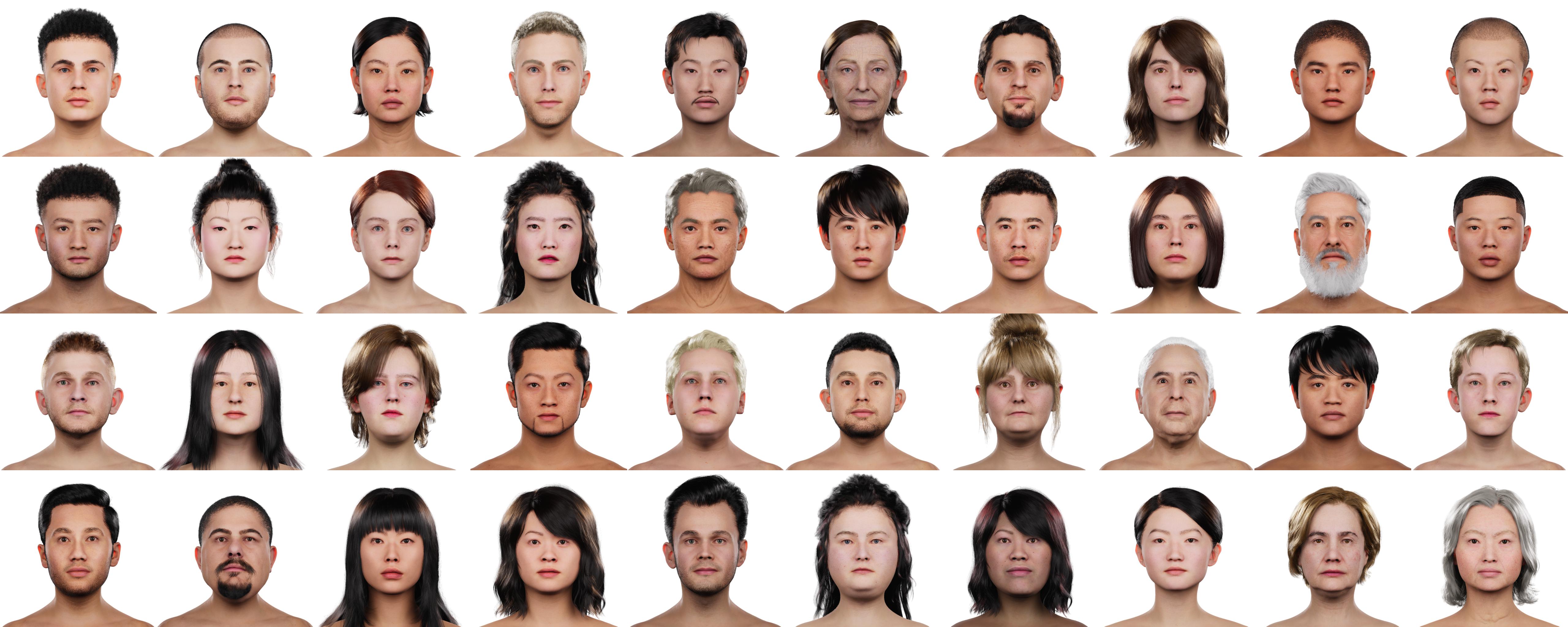}
    \vspace{-0.0in}
    \caption{We create a high-quality artist-designed 3D head dataset, containing 100 different subjects with various hairstyles
    }
    \vspace{-0.1in}
    \label{fig:supp_multi_id}
\end{figure*}

\begin{figure*}[!htbp]
    \centering
    \includegraphics[width=0.7\linewidth]{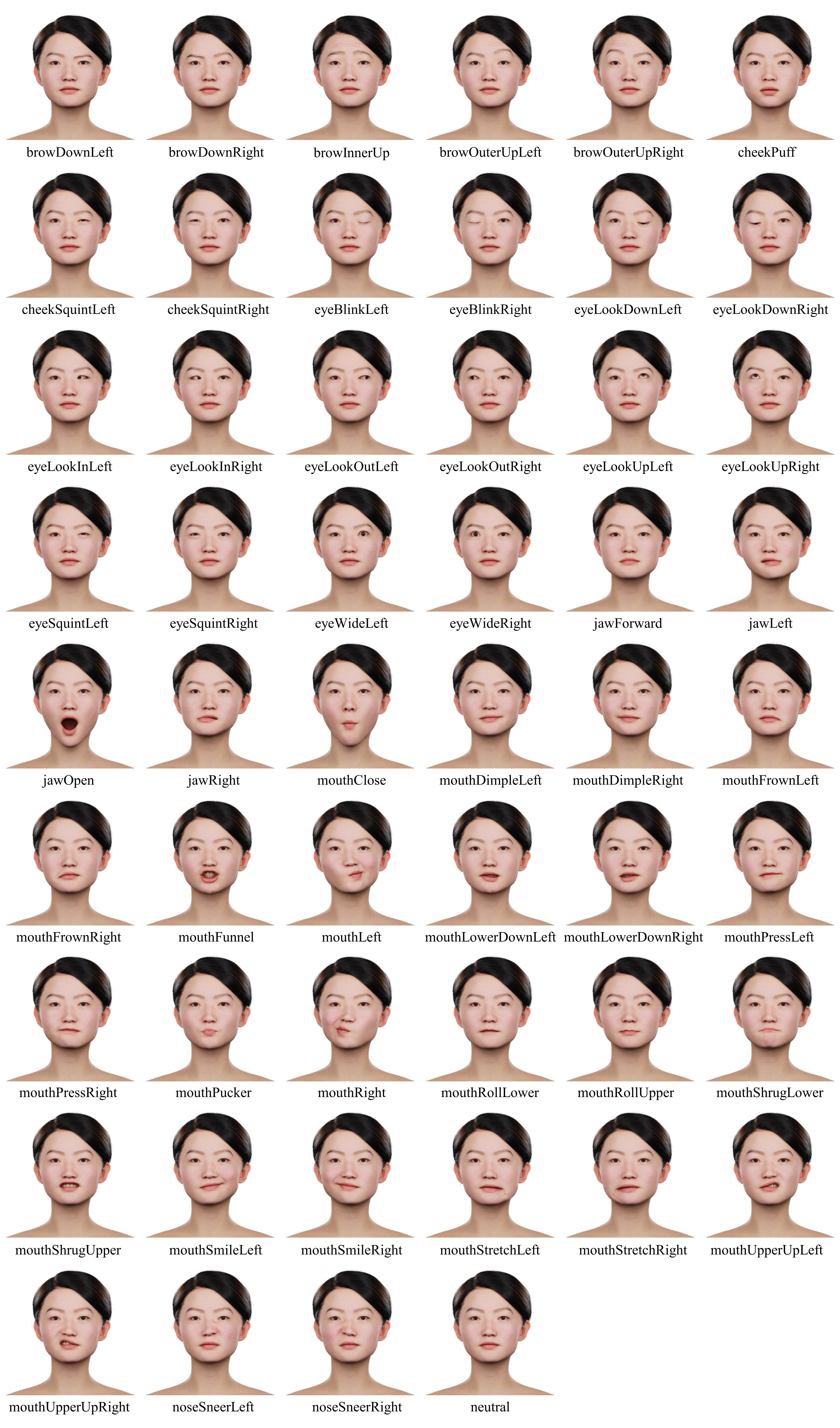}
    \vspace{-0.1in}
    \caption{
    The visualization of 52 blendshape bases we used.
    }
    \vspace{-0.1in}
    \label{fig:supp_blendshape}
\end{figure*}

We create a high-fidelity 3D head dataset for research use, containing 100 subjects, and each of them is rigged into 52 blendshapes bases. The 3D heads are rendered by 72 head-centric virtual cameras that cover 3 pitch angles and 24 horizontal rotation angles. Fig.~\ref{fig:supp_multiview} shows a subset of views of a single subject with one expression from our dataset. Fig.~\ref{fig:supp_blendshape} showcases the 52 standard expressions generated in our dataset, consistent with the blendshapes bases defined by ARKit~\cite{arkit}. Fig.~\ref{fig:supp_multi_id} presents more identities in our dataset.

\begin{figure*}[!htbp]
    \centering
    \includegraphics[width=0.95\linewidth]{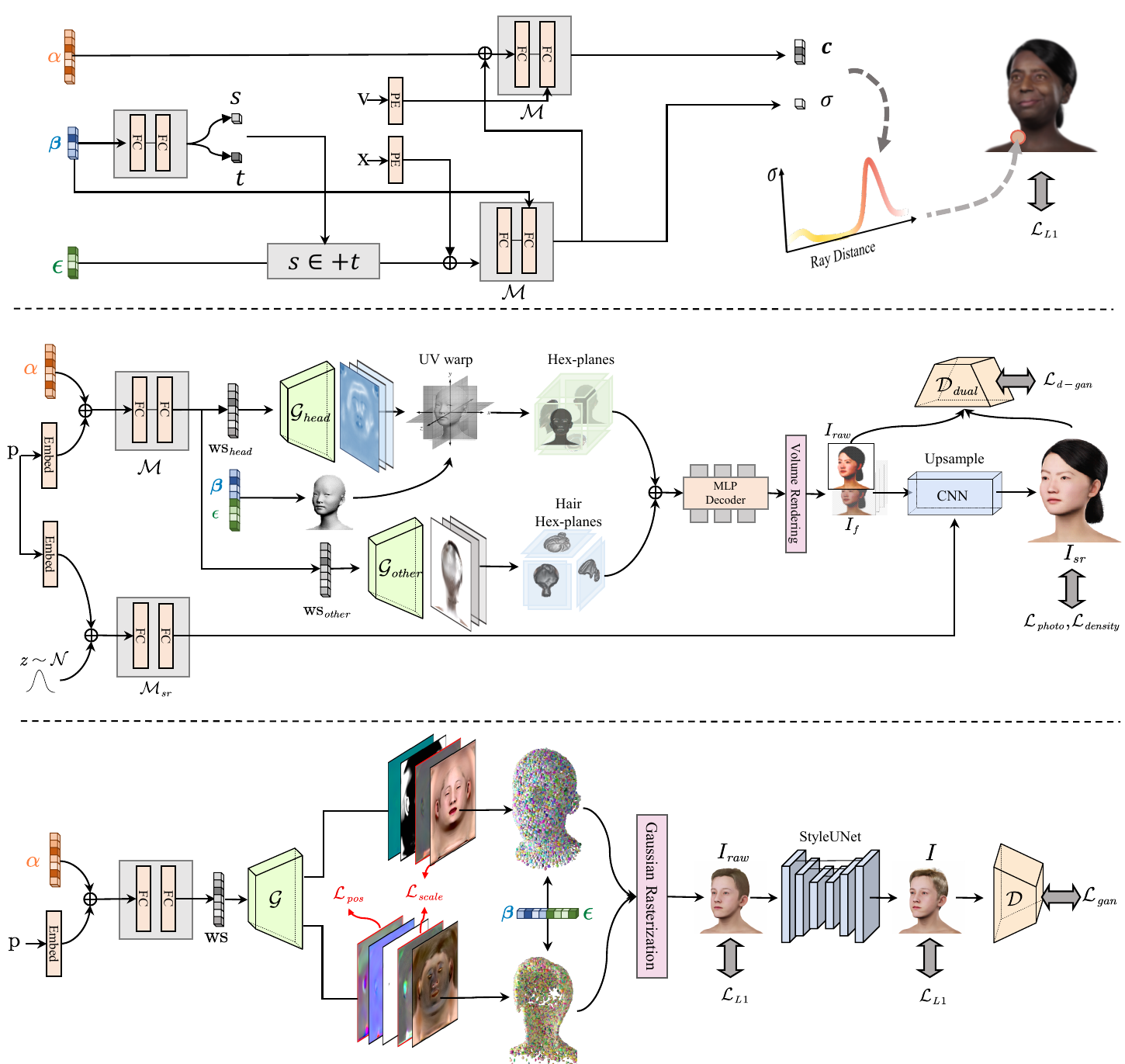}
    \vspace{-0.1in}
    \caption{
    The visualization of the three pipelines for different representations we used is shown from top to bottom: volume-based, hybrid-based, and point-based.
    }
    \label{fig:supp_archi}
\end{figure*}


\subsection{Implementation Details.}

\subsubsection{Volume-based}
Volume-based model is trained for roughly $400k$ iterations, which takes about 2 days on 4 NVIDIA GTX3090 GPUs.
In volume rendering, $1024$ rays are sampled in an iteration, each with $64$ sampled points in the coarse volume and additional $64$ in the fine volume.  The strategy of hierarchical volume sampling is used to refine coarse volume and fine volume simultaneously, as defined in NeRF\cite{mildenhall2020nerf}. The resolution of the images rendered by MoFaNeRF is $256\time256$, and the RefineNet takes the image rescaled to $512\times512$ as input, and synthesizes the final image in $512\times512$.
We use the Adam optimizer\cite{kingma2015adam} with the initial learning rate as $5\times{10^{-4}}$, decayed exponentially as $2\times{10^{-5}}$, $\beta_1=0.9$, $\beta_2=0.999$,  $\epsilon=10^{-7}$. 

\subsubsection{Hybrid-based}
Hybrid-based model is trained on 4 NVIDIA 3090 GPUs for roughly 10 days with batch size 16. The learning rates of the generator and the discriminator are 0.0025 and 0.002 respectively. The weight of R1 gradient regularization for discriminator is set to 4. To achieve a balance between training speed and rendering quality, the images in our dataset are downscaled to a resolution of $512\times512$ before being fed into the network, and the rendering structure of the model also outputs images at a resolution of $512\times512$. For hair swapping, dual hex-planes are generated for head and hair separately. The full network with two hex-planes trainable is firstly trained with the images with hairs for $3000$k iterations. Then the model is fine-tuned for hair detachment for additional $40$k iterations, where only hex-planes for hair are trainable. The hyperparameters of the optimizer are unchanged in this two-phase training.

In the task of facial fitting, we employ Poisson blending for facial swapping as part of our image preprocessing pipeline. To address occasional color inconsistencies along the facial boundary, we apply a gradient transition technique. Specifically, we extract a facial segmentation mask, which was eroded 20 times inward to obtain a new mask. We then perform a smooth transition from 0 to 1 in the non-overlapping regions between the two masks, followed by facial swapping in the RGB domain.

\subsubsection{Point-based}
Point-based model is trained on 8 NVIDIA A6000 GPUs for 2.5 days with 32 batch size. And learning rates are set to 0.02 for both generator and discriminator. $\gamma$ in R1 regularization is 5. During training, we decouple hair and head by sampling bald ground truth with a probability of 0.1. We end up training with 3200k iterations. In Gaussian rasterization, we directly render a $512\times512$ image. Additional details are synthesized using an appended StyleUNet, which ultimately outputs a result at the same resolution of $512\times512$.

\subsection{Details of Network}
We visualized the detailed frameworks of our models, as illustrated in Fig. \ref{fig:supp_archi}.
\subsubsection{RefineNet.}
A conditional GAN is introduced to improve the synthesized details further. Following MoFaNeRF~\cite{zhuang2022mofanerf}, we adopt pix2pixHD~\cite{wang2018high} as the backbone of our RefineNet. The input of the RefineNet is the generated images, which are rendered from the composited hex-planes. In the training phase, the overall network except for RefineNet is first trained, then RefineNet is trained with the other parts of the network detached. The loss function for training RefineNet is formulated as follows:

\begin{equation}
    \mathcal{L}_{refine} = \mathcal{L}_{gan} + \mu_{1} \mathcal{L}_{ssim} + \mu_{2} \mathcal{L}_{lpips}
\end{equation}

\noindent where $\mathcal{L}_{gan}$, $\mathcal{L}_{ssim}$, and $\mathcal{L}_{lpips}$ are GAN loss, SSIM loss, and perceptual loss, respectively. $\mu_{1}$ and $\mu_{2}$ are corresponding loss weights, which are set according to the settings of pix2pixHD~\cite{wang2018high}.

\subsubsection{{Volume-based}}
The volume-based model accepts input including the sampled points, view directions, and the parameters ($\alpha,\beta,\epsilon$). Initially, the sampled points are processed with the position encoding, then input through a $D$-layer MLP with a width of $W$, generating encoded spatial coordinates referred to as encoded coordinates. Then, these encoded coordinates are concatenated with the input body model codes and passed into another $D$-layer MLP with a width of $W$, producing shape features. The shape features then go through a linear layer to obtain depth values, representing depth information;
Next, the shape features are concatenated with the input UV codes and passed into another $D$-layer MLP with a width of $W$, yielding texture features. Subsequently, these texture features are concatenated with the input viewpoints and processed through a linear layer with a width of $W/2$, yielding an updated view dependent features; 
Finally, the view-dependent features pass through a linear layer to obtain color values, representing RGB color information. The shape features are decoded through a two-layer MLP to estimate the density value.

To enhance rendering results and representation ability, we follow a coarse-to-fine sampling strategy. We define an MLP with $D=256$ as the coarse network for importance sampling and an MLP with $D=1024$  as the fine network for the final output. The width $W$ is set to $8$ in both networks. 

\subsubsection{Hybrid-based}

We implement hybrid-based framework on top of the official Pytorch implementation of Next3D~\cite{sun2023next3d}.
To enhance the network's representation capability and representation capability, we project a volume onto six planes, corresponding to the positive and negative directions of the X, Y, and Z axes ($\pm{x}, \pm{y}, \pm{z}$). This method provides a more comprehensive representation of the scene compared to using only three planes. We then combine each pair of opposite-facing planes using an addition operation. Furthermore, we increase the feature dimension of each plane to 64 dimensions.

The model receives a 50-dimensional identity encoding to the generative process. The identity encoding is concatenated with the embeded camera pose and then projected into the latent space by the 2-layer MLP. And texture generator progressively synthesizes features from low to high resolution, using modulated convolutional layers that upsample feature maps from $4\times4$ to $256\times 256$, with several intermediate resolutions ($8\times8,16\times16$, etc.). Hair and head are modeled by two separate texture generators. When hex-planes and tri-planes are incorporated, the network processes rays through ray samplers to obtain 3D coordinates for rendering. Following this step, the model can produce a raw image with a resolution of $64\times64$. Ultimately, by incorporating a super-resolution module, the model enhances the resolution of the raw image to $512\times512$. In particular, we apply a separate mapping network at the super-resolution stage. It maps gaussian noise into the latent space, which is used to modulate the super-resolution module. Empirical study indicates that it helps in enhancing the quality of the hair.


\subsubsection{Point-based}

The point-based model utilizes a mapping network and generator that are essentially the same as those in the hybrid-based model. However, the generator in the point-based model produces more specific outputs. It generates feature maps with a total of 26 channels—13 for the head and 13 for the hair. For each set of 13 channels, 3 represent RGB, 3 are for Gaussian scale, 3 for Gaussian rotation (in axis-angle representation), 1 for opacity, and 3 for the offset of the Gaussian position. The offset for the head is always zero, as we already have an accurate mesh, while the offset for the hair is constrained using a tanh activation to ensure it stays within a reasonable range. Following GGHead, we initialize the offset channels to zero via a zero convolution operation and map the scale texture to a reasonable range using a softplus-based nonlinear transformation. Specifically, in the StyleUNet stage, we set the modulation vector to a fixed dummy variable. We find that using the embedding of the identity code in the latent space can lead to difficulties in decoupling.


\section{More Experimental Results}
\label{sec:more_exp}

 \begin{figure}[th]
    \centering
    \includegraphics[width=0.75\linewidth]{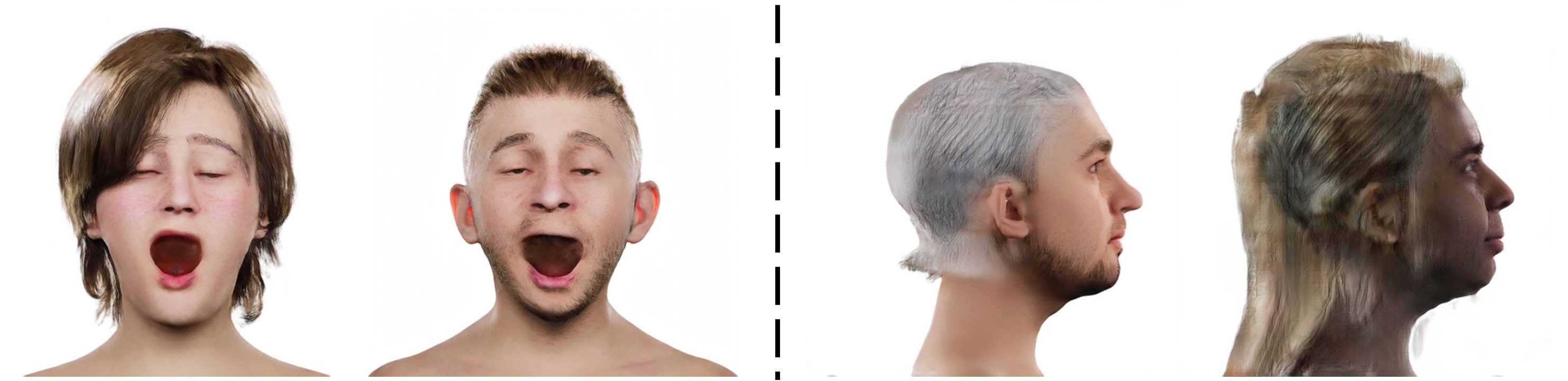}
    \vspace{-0.0in}
    \caption{Failure cases about the wide opening mouth and specific hairstyles. 
    }
    \vspace{-0.1in}
    \label{fig:supp_failure}
\end{figure}

 \begin{figure}[th]
    \centering
    \includegraphics[width=1.0\linewidth]{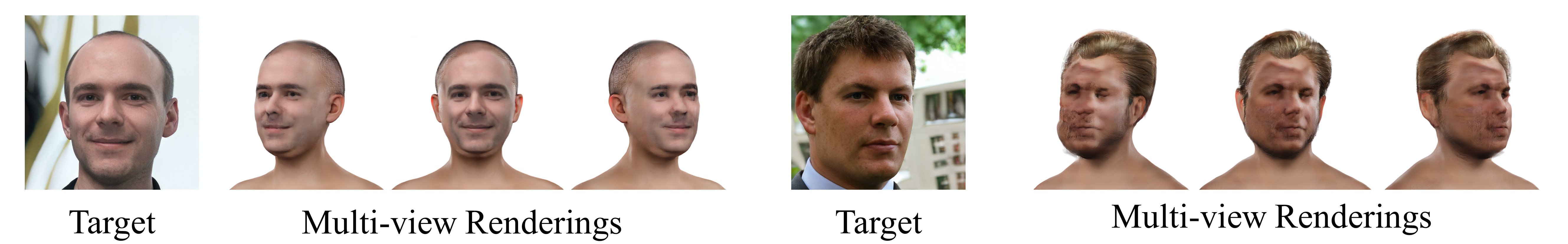}
    \vspace{-0.2in}
    \caption{Bad cases in image-based fitting. 
    }
    \vspace{-0.1in}
    \label{fig:supp_failfit}
\end{figure}

\subsection{Interpolation in the Parametric Space}
Our parametric model decomposes the full head into three distinct parameters: the appearance code $\alpha$, shape code $\beta$, and expression code $\epsilon$. This decomposition allows for the generation of reasonable results when interpolation is applied to these defined parameters. 

More importantly, the results demonstrate that each parameter makes a specific impact on the results, indicating a successful decomposition process. 
For a dynamic representation of these interpolation results, please refer to the supplemental video.

\subsection{More Visual Results}
\label{sec:more_res}
We provide the results of the animation and 360-degree rendering in the supplementary video. This demonstrates that our parametric full-head model is capable of detaching the hair without impacting the primary functionality of our method.

\subsection{Comparison with a few face models}
\begin{figure}[!htbp]
    \centering
    \includegraphics[width=1\linewidth]{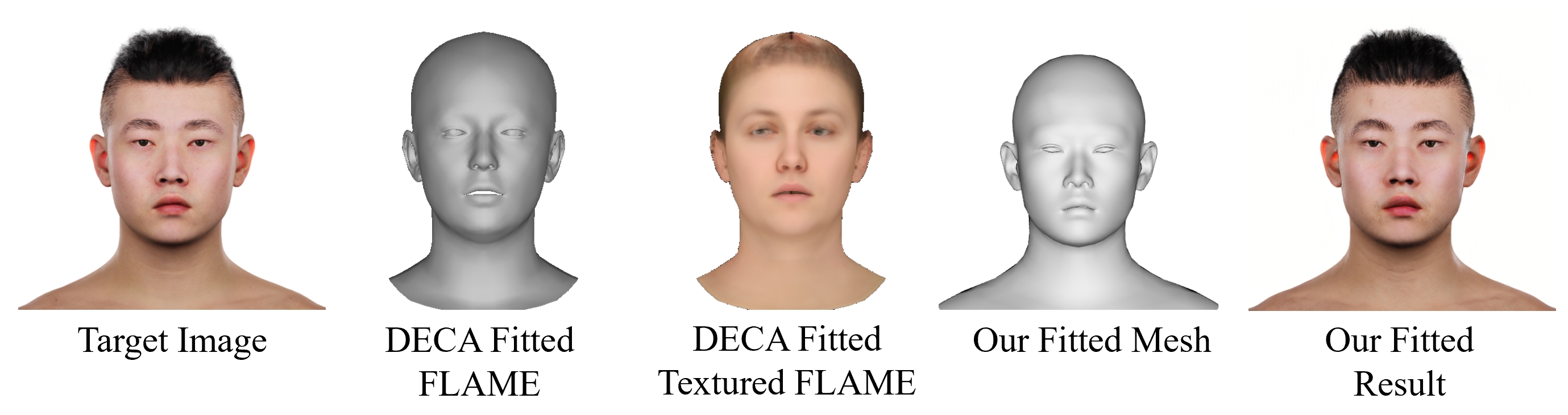}
    \vspace{-0.1in}
    \caption{
    Comparison with FLAME model.
    }
    \vspace{-0.3in}
    \label{fig:flame_comp}
\end{figure}

FLAME\cite{li2017learning} and JNR\cite{JNR} are both parametric mesh models for 3D head representation, which struggle to represent the inner mouth, eye-balls, and hair for the complex material of these regions. By contrast, our artist-designed models are delicately rigged with high-fidelity hair, eyeballs and mouth cavity. A comparison with FLAME fitted results is shown in Fig.~\ref{fig:flame_comp}.

\subsection{Limitations and Bad Cases}
\label{sec:limi}

There are still some limitations to our approach. Firstly, the stability of single-image fitting still needs to be improved, and the fitting method only works when the target face is frontal. If the target has a very unusual hairstyle, the fitting method may fail to produce a plausible appearance. Secondly, we found that details in the oral cavity, such as teeth and tongue, are lost when the mouth is wide open, as shown on the left side of Fig.~\ref{fig:supp_failure}. When changing the hair color from black hair to specific light colors, a degraded texture and artifacts may exist, as shown on the right of Fig.~\ref{fig:supp_failure}.  
In the task of image-based fitting, the performance may be degraded when the target face is turned aside, as shown in Fig.~\ref{fig:supp_failfit}.
We will explore to address these issues in the future work.

\end{document}